\pdfoutput=1

\documentclass[11pt]{article}

\usepackage{EMNLP2023}

\usepackage{times}
\usepackage{latexsym}

\usepackage[T1]{fontenc}

\usepackage[utf8]{inputenc}

\usepackage{microtype}

\usepackage{inconsolata}

\usepackage{graphicx} 
\usepackage[T1]{fontenc}    
\usepackage{hyperref}       
\usepackage[capitalize]{cleveref}
\usepackage{url}            
\usepackage{booktabs}       
\usepackage{amsfonts}       
\usepackage{nicefrac}       
\usepackage{xcolor}         
\usepackage{multirow}
\usepackage{tabularx}
\usepackage{caption}
\usepackage{graphicx,lipsum}

\begin{document}

\title{Auto-Evolve: Enhancing Large Language Model's Performance via Self-Reasoning Framework}

\author{%
    Krishna Aswani \footnotemark[1] \quad Huilin Lu \footnotemark[1] \quad Pranav Patankar \footnotemark[2] 
    \\
    {\bf Priya Dhalwani  \quad Iris Tan \quad
    Jayant Ganeshmohan  \quad Simon Lacasse } \vspace{1ex}
    \\
    Amazon
}

\maketitle

\renewcommand{\thefootnote}{\fnsymbol{footnote}}
\footnotetext[1]{Equal contribution}
\footnotetext[2]{Corresponding Author: \href{mailto:pppatan[at]amazon[dot]com}{pppatan [at] amazon [.] com}}

\begin{abstract}
Recent advancements in prompt engineering strategies, such as Chain-of-Thought (CoT) and Self-Discover, have demonstrated significant potential in improving the reasoning abilities of Large Language Models (LLMs). However, these state-of-the-art (SOTA) prompting strategies rely on single or fixed set of static seed reasoning modules like \emph{"think step by step"} or \emph{"break down this problem"} intended to simulate human approach to problem-solving. This constraint limits the flexibility of models in tackling diverse problems effectively. In this paper, we introduce Auto-Evolve, a novel framework that enables LLMs to self-create dynamic reasoning modules and downstream action plan, resulting in significant improvements over current SOTA methods. We evaluate Auto-Evolve on the challenging BigBench-Hard (BBH) dataset with Claude 2.0, Claude 3 Sonnet, Mistral Large, and GPT 4, where it consistently outperforms the SOTA prompt strategies. Auto-Evolve outperforms CoT by up to 10.4\% and on an average by 7\% across these four models. Our framework introduces two innovations: a) Auto-Evolve dynamically generates reasoning modules for each task while aligning with human reasoning paradigm, thus eliminating the need for predefined templates. b) We introduce an iterative refinement component, that incrementally refines instruction guidance for LLMs and helps boost performance by average 2.8\% compared to doing it in a single step. 
\end{abstract}

\section{Introduction}

LLMs have demonstrated significant potential in  various Natural Language Processing (NLP) capabilities such as understanding, generating, and reasoning \citep{brown2020language,chowdhery2022palm, anil2023palm, openai2023gpt}. Despite the impressive progress, LLMs continue to have challenges in solving multi-step reasoning tasks that require systematic thinking and planning. Increasing model size alone is not enough to solve these issues, emphasizing the necessity for developing novel techniques to improve LLMs' reasoning capabilities \citep{srivastava2023beyond,DBLP:journals/corr/abs-2112-11446} .

Various prompting strategies have been developed to guide and facilitate the reasoning capabilities of LLMs. CoT \citep{wei2022chain} has emerged as a prominent approach, encouraging LLMs to generate step-by-step explanations mimicking human reasoning. 

Subsequent research efforts have focused on refining the generation process thereby enhancing the quality and consistency of the rationales \citep{kojima2022large,fu2023complexitybased,zhou2022least,wang2022self}. 
 Self-Discover \citep{zhou2024selfdiscover} improves models' reasoning capabilities over CoT by allowing models to select the most appropriate reasoning path from a fixed set of reasoning modules. However, our analysis of the seed modules in Self-Discover revealed that a subset of fixed seed modules dominated the usage, limiting the framework's reasoning coverage and performance on diverse tasks (Appendix: \cref{fig:seed_modules_analysis}). CoT's and Self-Discover's reliance on a limited set of reasoning seed modules such as "think step by step" or "break down this problem" constrains the approaches to tackling a problem, negatively affecting their ability to generalize over diverse tasks. 
\begin{figure*}
  \centering
  \includegraphics[width=\textwidth,height=7cm]{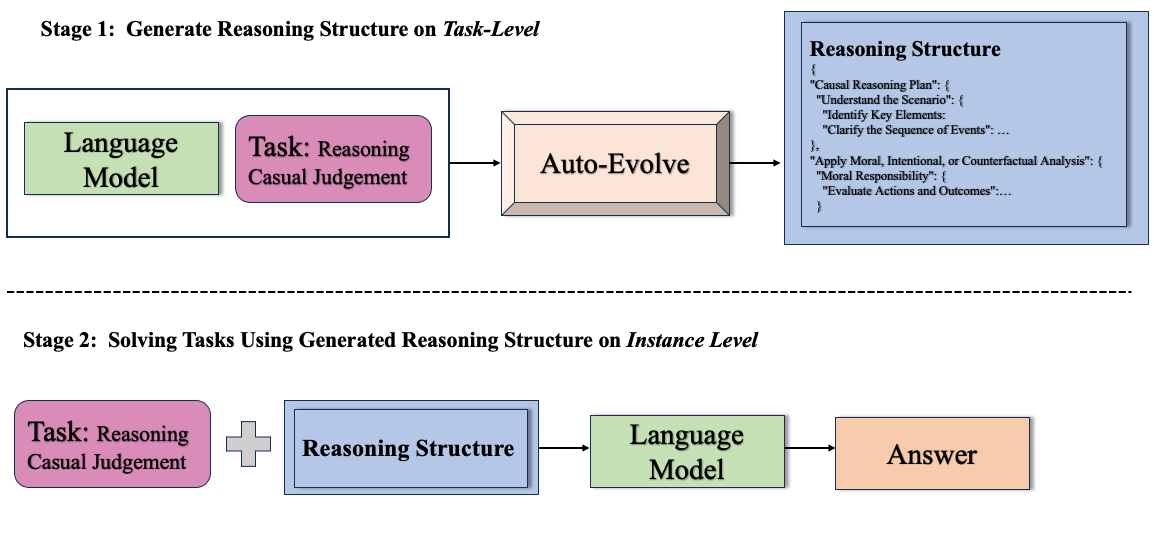}
  \caption{Illustration of using Auto-Evolve workflow for problem-solving. }
  \label{fig:self-evolve two stage}
\end{figure*}

Our framework, Auto-Evolve, builds upon the strengths of previous prompting approaches while addressing their limitations. Rather than relying on a fixed set of seed modules, Auto-Evolve creates custom reasoning modules on-the-fly for each task, allowing LLMs to come up with a wider range of reasoning structures (instruction guidance in JSON format that LLMs can follow to solve a task step by step) that are better suited to handle the specific needs of each task. Auto-Evolve further incorporates an iterative refinement process allowing models to refine their reasoning structures based on the specific requirements of each task, significantly boosting performance. 

Auto-Evolve consists of three core components: 

\textbf{1) Reasoning Module Generator} that  dynamically creates relevant modules for a given task.

\textbf{2) Reasoning Structure Initializer} that composes an initial reasoning plan using the generated modules.

\textbf{3) Reasoning Structure Evolver} that iteratively refines and improves the plan over multiple processing steps.

We evaluate Auto-Evolve's performance on Big Bench Hard \citep{suzgun2022challenging}, a widely used
benchmark with a subset of 23 hard tasks from the
BIG-Bench suite \citep{srivastava2023beyond}. On average, across four models, Auto-Evolve demonstrates a \textbf{12.8\%} performance improvement over Direct Prompt (Claude 2.0: 11.7\%, Claude 3 Sonnet: 3.0\%, Mistral Large: 13.5\%, GPT-4: 22.9\%), a \textbf{7\%} performance improvement over CoT  (Claude 2.0: 10.4\%, Claude 3 Sonnet: 3.3\%, Mistral Large: 8.1\%, GPT-4: 6.3\%) and a \textbf{4\%} improvement over Self-Discover \textbf{4\%} (Claude 2.0: 6.7\%, Claude 3 Sonnet: 3.9\%, Mistral Large: 2.7\%, GPT-4: 2.6\%). By combining dynamic prompt generation and iterative refinement, Auto-Evolve offers a more flexible and adaptive approach to reasoning, pushing the boundaries of LLMs performance on complex tasks.

\section{Related work} 

There are two key model optimization techniques, Model Prompting and Model Fine-Tuning. Model Prompting Methods enhance the reasoning capabilities of LLMs by providing carefully designed prompts that guide the model towards generating the desired output, without modifying the underlying model parameters. On the other hand, Model Fine-Tuning Methods involve updating the model's parameters by training on a relevant dataset to specialize the model for a particular task or domain, which can be computationally expensive.
While both methods have their advantages and disadvantages, our framework Auto-Evolve, is closely related to Model Prompting Methods.


\textbf{Model Prompting Methods} such as CoT prompting \citep{wei2022chain} encourages models to generate intermediate reasoning steps that lead to the final desired answer. CoT has been shown to boost performance on arithmetic, commonsense, and symbolic reasoning tasks. Subsequent work has extended CoT by selectively sampling rationales \citep{kojima2022large}, improving rationale consistency (\citet{wang2022self}; Self-Consistency), generating more structured reasoning paths \citep{fu2023complexitybased}, and having models first plan the reasoning before solving the problem (\citet{wang2023planandsolve}; Plan-and-Solve). Self-Discover \citep{zhou2024selfdiscover}, introduces a three-stage process where LLMs select relevant reasoning modules, adapt them to the specific task, and implement them into a coherent reasoning structure. Self-Discover outperforms CoT \citep{wei2022chain},  Self-Consistency \citep{wang2022self} and Plan-and-Solve \citep{wang2023planandsolve} prompting on various benchmarks.

\begin{figure*}[h]
  \centering
  \includegraphics[width=\textwidth,height=7cm]{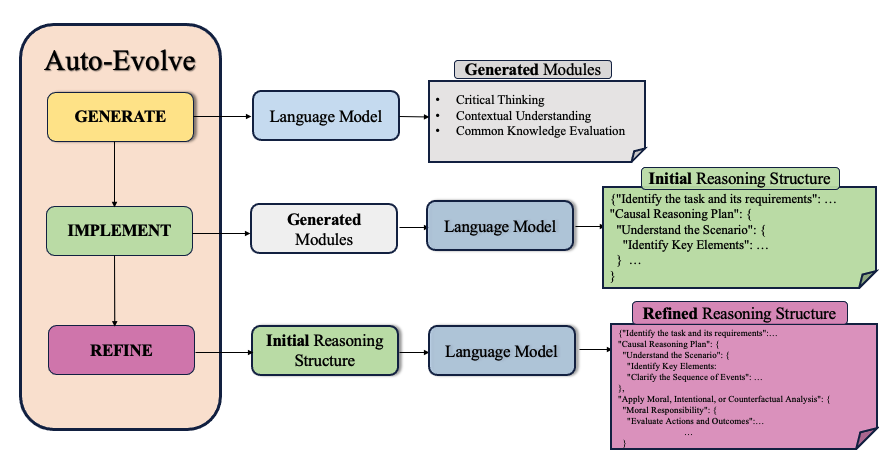}
  \caption{Overview of three components of Auto-Evolve Stage 1. Component Reasoning Module Generator \textbf{GENERATE} a set of task-specific reasoning modules and component Reasoning Structure Initializer \textbf{IMPLEMENT} a starting JSON reasoning structure. 
  Over multiple runs of \textbf{REFINE}, component Reasoning Structure Evolver subsequently refines the reasoning structure to a domain-adaptive actionable plan.For instance, when solving the reasoning QA task, the initial reasoning structure from \textbf{IMPLEMENT} may lack depth in 'moral, intentional, or counterfactual analysis'. The \textbf{REFINE} process addresses this gap by identifying and incorporating these additional elements, thus improving the structure's ability to solve the task.}
  \label{fig:self-evolve framework}
\end{figure*}

\section{Auto-Evolve Framework}
\begin{figure*}
    \centering
    \begin{minipage}{0.48\textwidth}
        \centering
        \includegraphics[width=\linewidth]{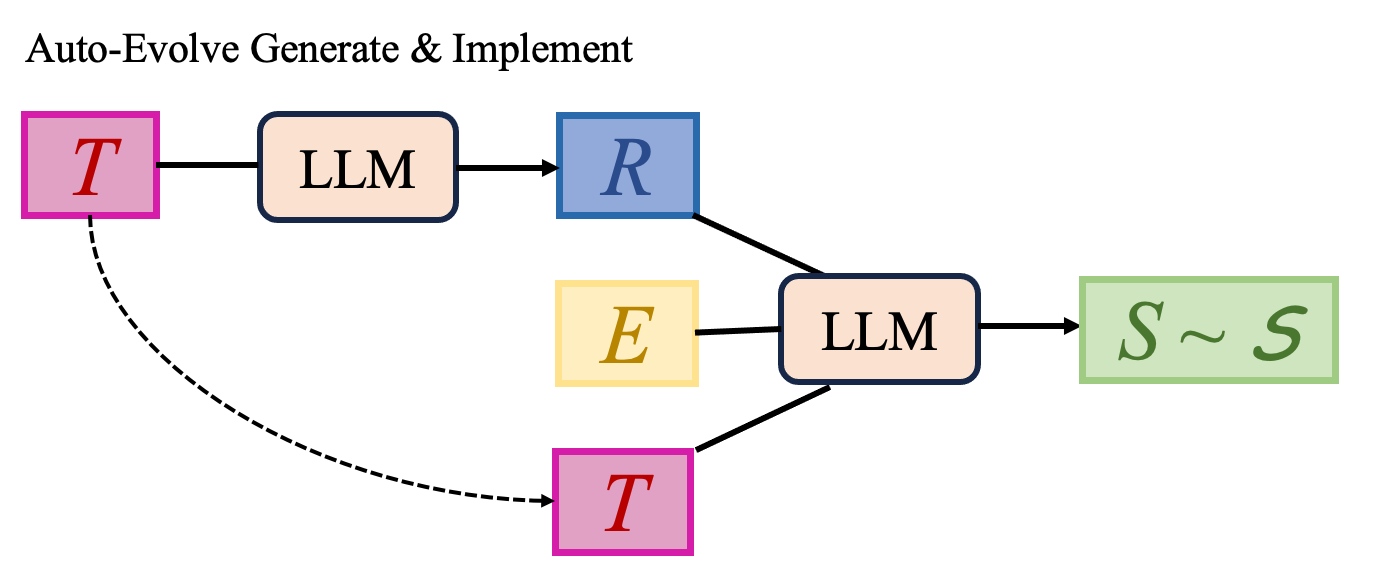}
    \end{minipage}\hfill
    \begin{minipage}{0.45\textwidth}
        \centering
        \includegraphics[width=\linewidth]{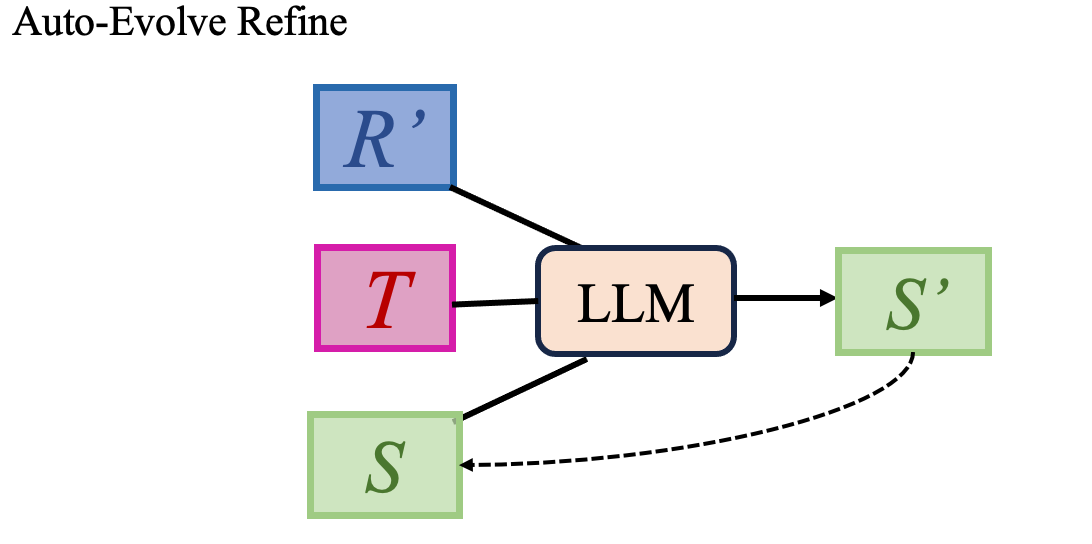}
    \end{minipage}
    \caption{Overview of Auto-Evolve workflow in mathematical notation }
    \label{fig:multiple-variants}
\end{figure*}

Auto-Evolve framework is inspired by two fundamental principles. (1) \textbf{Higher interpretability associated with JSON structure}: LLM's reasoning capabilities and performance are enhanced by JSON structure's higher interpretability \citep{zhou2023far,openai2023gpt,openai_json}. (2) \textbf{LLMs have inbuilt diverse reasoning abilities}: LLMs possess an inherent grasp of diverse thinking styles and essential reasoning modules crucial for tackling variety of tasks since they were trained on enormous data, typically measured in petabytes. SOTA Self-Discover adheres to the first principle, but it overlooks the key aspect of the second principle. Instead of leverage knowledge hidden within LLMs, Self-Discover supplies LLMs with a fixed set of initial human-designed reasoning modules such as \emph{“Use critical thinking”} and \emph{“Let’s think step by step”}. On the flip side, Auto-Evolve advocates for LLMs' intrinsic ability to independently discern and utilize relevant reasoning strategies for different tasks. 
\newline
Auto-Evolve comprises of two stages as illustrated in \cref{fig:self-evolve two stage}. Stage 1 dynamically generates intrinsic task-related reasoning modules and structure (JSON instructions) by leveraging task examples and three meta-prompts, thereby guiding LLMs to solve tasks without needing static human-designed seed modules and further training. Stage 1 operates at \emph{task-level}, i.e., one run for each task category. Stage 2 uses the finalized reasoning structure produced as an output of Stage 1 to solve individual \emph{task instances} by asking the model to follow the instruction step by step. 
Given the straightforward and uncomplicated nature of Stage 2, we focus rest of this section on further elaborating the three components of Stage 1 that are illustrated in \cref{fig:self-evolve framework}. We also present a graphical representation in \cref{fig:multiple-variants} that's accompanied by mathematical notations to elucidate the procedure of Stage 1. Left half of this figure showcases the \textbf{GENERATE} and \textbf{IMPLEMENT} components, while right half showcases the \textbf{REFINE} components. The mathematical notations are explained in the following subsections. Prompts details are included in Appendix \cref{fig:self-evolve promt detail}.

\subsection{Reasoning Module Generator (GENERATE)}
The primary function of the \textbf{GENERATE} component is to dynamically create task-specific reasoning modules and descriptions. Unlike the Self-Discover approach that relies on a predetermined set of 39 static reasoning modules (Appendix \cref{fig:seed_modules_analysis}) for problem solving, \textbf{GENERATE} embraces adaptability and responsiveness by creating modules dynamically. E.g., the reasoning modules in Appendix \cref{fig:deep_dive_reasoning_module} for Boolean Expression and Disambiguation QA tasks are generated using Auto-Evolve.

For the tasks under the same domain, given only a few task examples without labels $t_i\in T$, \textbf{GENERATE} first creates a set of task-specific reasoning modules $\mathcal{R}$ by using a model $\mathcal{M}$ and a meta-prompt $\mathcal{P}_G$:
\begin{equation}
\mathcal{R} = \mathcal{M}(\mathcal{P}_G || t_i). 
\end{equation}
By assessing the unique attributes and demands of each task, \textbf{GENERATE} orchestrates the creation of a task-specific set of reasoning modules, ensuring a nuanced and tailored approach to problem-solving. The dynamic generation process enables our framework to continually evolve and adapt to new challenges and task domains, facilitating more effective and contextually relevant reasoning processes.

\subsection{Reasoning structure initializer (IMPLEMENT)}
\textbf{IMPLEMENT} serves as a starting point for generating task-specific reasoning structure. \textbf{IMPLEMENT} uses only the first reasoning module from \textbf{GENERATE} for building the initial reasoning structure. This lays the groundwork for subsequent refinement steps and ensures the initial reasoning structure aligns closely with the context of the given task.

Given the same task examples without labels $t_i\in T$, Reasoning Structure Initializer implements an initial \emph{key-value} reasoning plan $\mathcal{S}$  by using the first reasoning module $\mathcal{R}_1$ generated from previous component, an action plan of another task $E$ and a meta-prompt $\mathcal{P}_I$:
\begin{equation}
\mathcal{S} = \mathcal{M}(\mathcal{P}_I || t_i || \mathcal{R}_1 || E). 
\end{equation}

\subsection{Reasoning structure evolver (REFINE)}
Finally, given the initial reasoning structure $\mathcal{S}$, \textbf{REFINE} component iteratively distills the initial reasoning structures by incorporating additional reasoning modules $\mathcal{R}_i$ generated by the Reasoning Module Generator. This component also uses a meta-prompt $\mathcal{P}_E$, an example-agnostic structured prompt designed to capture the reasoning structure of a specific category of tasks. During the iterative refine process, the generated new reasoning structure $\mathcal{S'}$ will replace the original $\mathcal{S}$ and be used for the next iteration. By dynamically evolving the reasoning structure in this manner, our approach fosters a comprehensive and versatile framework capable of addressing a wider range of cognitive tasks. Through empirical evaluation, we demonstrate the efficacy of our methodology in improving reasoning performance and adaptability across various task domains, thereby contributing to enhancing the language models reasoning capabilities.
\begin{equation}
\mathcal{S'} = \mathcal{M}(\mathcal{P}_E || \mathcal{R}_i || \mathcal{S}). 
\end{equation}

\section{Experiments} 
\subsection{Datasets}

We evaluate Auto-Evolve using a diverse and large-scale reasoning benchmarking dataset: BIG Bench Hard (BBH) \citep{suzgun2022challenging}. It is designed to evaluate the performance and reasoning capabilities of language models. It consists of 23 complex reasoning tasks, totaling 5,511 task instances. (Appendix \cref{table:bbh_pertask}) spanning across 4 domains: (1) Algorithmic and Multi-Step Arithmetic Reasoning (11 tasks, e.g., \emph{Boolean Expressions Evaluation}, \emph{Object Counting}), (2) Natural Language Understanding (7 tasks, e.g., \emph{Snarks}, \emph{Disambiguation QA}), (3) Use of World Knowledge (5 tasks, e.g., \emph{Movie Recommendation}, \emph{Date Understanding}), and (4) Multilingual Knowledge and Reasoning (\emph{Salient Translation}). We use accuracy as the evaluation metric to measure the model performance on BBH.

\subsection{Models}
We use four LLMs to showcase the generalizability of Auto-Evolve framework: Claude 2.0 \citep{claude-2.0}, Claude 3 Sonnet \citep{claude-3.0}, Mistral Large \citep{mistral-l} and GPT-4 (gpt-4-turbo-preview) \citep{openai2023gpt}. In our experiments, LLMs exhibited non-determinism even with temperature set to 0\footnote{The Non-Determinism of OpenAI and Anthropic Models - https://standardscaler.com/2024/03/06/the-non-determinism-of-openai-and-anthropic-models/}. To ensure robustness in our evaluations, we run all experiments three times and average the results. \cref{table:bbh_avg_abs}, \cref{table:bbh_avg}, \cref{fig:acc_diff_mistral} and \cref{fig:categories_performance} in the next section show the performance of Auto-Evolve compared to other prompt strategies.

\subsection{Baselines}
We compare Auto-Evolve with Direct, CoT and Self-Discover frameworks for evaluating LLM reasoning capabilities:
\newline\textbf{Direct Prompting}, where language models produce the answer without the need for intermediate reasoning stages.
\newline\textbf{CoT} \citep{wei2023chainofthought, kojima2022large}, where language models are prompted to produce a logical sequence of steps resulting in the final solution.
\newline\textbf{Self-Discover} \citep{zhou2024selfdiscover}, where a set of thinking styles are provided to guide LLMs to produce a logical path for solving problems, much like the approach a human expert might take.

\subsection{Experiments setup and evaluation}
\textbf{LLM Inputs:} For Direct Prompting, we only provide task instance as the prompt, while for CoT, we add an additional sentence \emph{"Thinking step-by-step"} to the prompt fed into the LLMs. For Self-Discover, the prompt includes a set of 39 thinking styles for LLMs to select and adapt to the tasks. For Auto-Evolve, however, we purely rely on LLMs to dynamically generate the task-specific reasoning modules and structures. During the steps for generating the task-specific reasoning modules and reasoning structures, we randomly select two task instances without the target labels from the task set as the examples fed to LLMs. For the step-by-step plan example which is applied in Reasoning Structure Initializer component of Stage 1, we use the model-discovered JSON structure generated from another task.

\textbf{LLM Response Evaluation:} We meticulously examine the results obtained from the LLMs with automatic and manual evaluation procedures. Since LLMs do not always produce consistent format of outputs when they follow the reasoning instructions, we programmatically extract answers/labels by examining the model responses. For the outputs that can not be programmatically extracted, we employ annotators to manually evaluate the model responses. 
Non-determinism in LLMs output means that slight variations in reasoning modules for both Self-Discover and Auto-Evolve lead to significant disparities in the downstream output of reasoning structures generation. Consequently, we experiment on all four LLMs for three times across all tasks, and calculate the average accuracy, ensuring robustness and fairness in our findings. This approach not only enhances the credibility of our results but also ensures consistency and validity in our experimental methodology.

\section{Results and Discussion}

\subsection{Performance}
\begin{table}[!b]
\caption{ Comparing absolute performances of Auto-Evolve against CoT \& Self-Discover prompting techniques. }
\medskip
\label{table:bbh_avg_abs}
\centering
\begin{tabular}{ lccc}
\toprule
\textbf{Method}&  \textbf{BBH}\\
\midrule
Claude 2.0 Direct& 53.7\%\\
Claude 2.0 + CoT&  55.0\%\\
Claude 2.0 + Self-Discover&  58.7\%\\
Claude 2.0 + Auto-Evolve&  \textbf{65.4\%}\\
\midrule
Claude 3 Sonnet Direct& 68.6\%\\
Claude 3 Sonnet + CoT&  68.3\%\\
Claude 3 Sonnet + Self-Discover&  67.7\%\\
Claude 3 Sonnet + Auto-Evolve&  \textbf{71.6\%}\\
\midrule
Mistral-Large Direct& 61.9\%\\
Mistral-Large + CoT&  67.3\%\\
Mistral-Large + Self-Discover&  72.7\%\\
Mistral-Large + Auto-Evolve&  \textbf{75.4\%}\\
 \bottomrule
\end{tabular}
\end{table}
Auto-Evolve demonstrates significant performance improvement across the 23 diverse tasks in the BBH dataset  \citep{suzgun2022challenging}. As shown in \cref{table:bbh_avg_abs},  Auto-Evolve achieves an average absolute \textbf{8.1\%} and \textbf{2.7\%} improvement across 23 diverse tasks over CoT and Self-Discover respectively when using Mistral Large. With Claude 2.0, the improvement is even more substantial, with \textbf{10.4\%} and \textbf{6.7\%} gains over CoT and Self-Discover. We observe the same trends for GPT-4 in \cref{table:bbh_avg}, where Auto-Evolve improves GPT-4's performance over CoT and Self-Discoverwith absolute gains of \textbf{6.3\%} and \textbf{2.6\%}. The performance improvements on Claude 3 Sonnet are less significant, achieving an average absolute \textbf{2.5\%} and \textbf{3.1\%} improvement over CoT and Self-Discover respectively. It is likely due to Claude 3 Sonnet's already advanced reasoning capabilities, which enable the model to perform exceptionally well even with a direct approach, without the aid of prompting techniques. This leaves less room for enhancement through external reasoning frameworks like Auto-Evolve. These results highlight the effectiveness of Auto-Evolve's dynamic and adaptive reasoning approach compared to frameworks that rely on static seed modules.

\begin{table}[h]
\caption{ Comparing delta performances across all tasks with Auto-Evolve against CoT \& Self-Discover for GPT-4. In table, it shows the absolute percentage improvement over baseline.}
\medskip
\label{table:bbh_avg}
\centering
\begin{tabular}{ lccc}
\toprule
\textbf{Method}&  \textbf{BBH}\\
\midrule
GPT-4 Direct (Baseline)& *\\
GPT-4 + CoT&  +16.6\%\\
GPT-4 + Self-Discover&  +20.3\%\\
GPT-4 + Auto-Evolve&  \textbf{+22.9\%}\\

 \bottomrule
\end{tabular}
\end{table}

 In \cref{fig:acc_diff_mistral} we highlight results from Mistral with other models results being available in Appendix \ref{Auto-Evolve Performance Comparison}. It provides a detailed breakdown of performance improvements across individual tasks. Auto-Evolve improves Mistral Large's performance over Self-Discover on \textbf{18}/23 tasks and surpasses CoT on \textbf{17}/23 tasks. We demonstrate that Auto-Evolve excels at tasks that require tracking complex problems such as Geometric Shapes, Web of Lies. The reasoning structures generated by Auto-Evolve assist LLMs in managing and solving these evolving problems. The dynamic generation of task-specific reasoning modules allows Auto-Evolve to effectively adapt to each unique challenge posed by individual tasks. Further task-level comparisons with other frameworks are available in the \cref{table:bbh_pertask}. 
 
\begin{figure*}[h]
    \centering
    \begin{minipage}{0.33\textwidth}
        \centering
        \includegraphics[height=5cm, width=\linewidth]{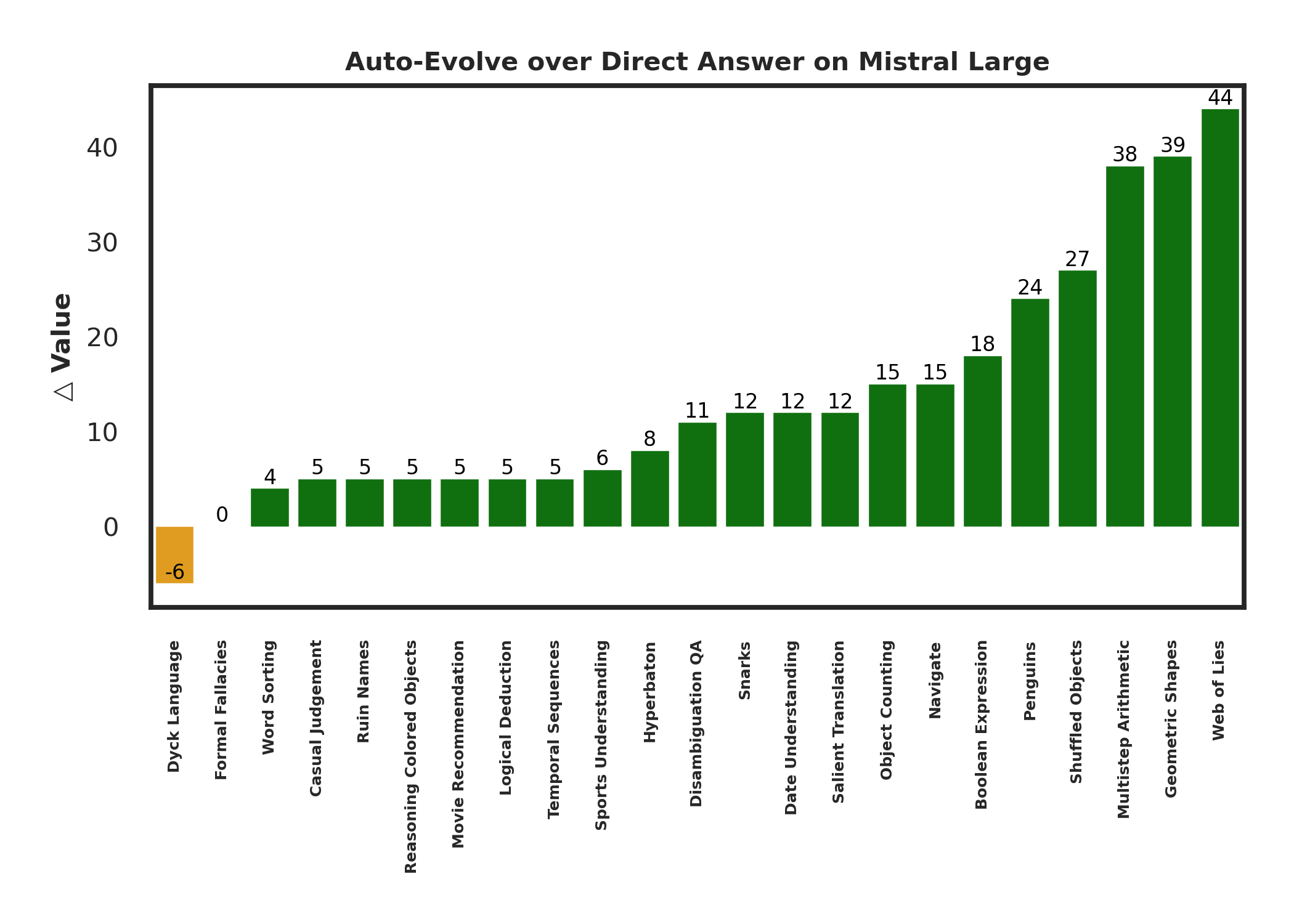}
    \end{minipage}
    \begin{minipage}{0.33\textwidth}
        \centering
        \includegraphics[width=\linewidth, height=5cm]{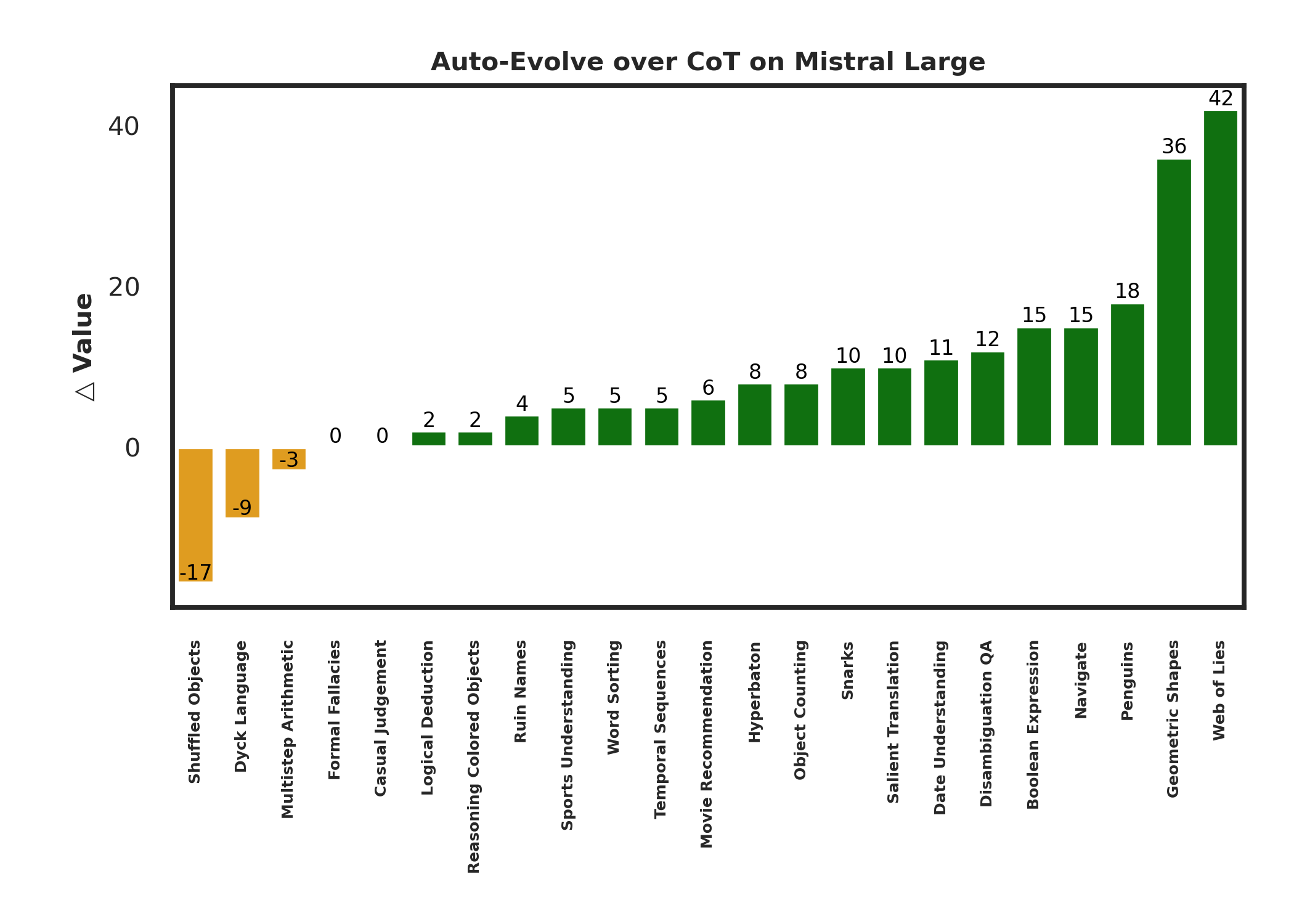}
    \end{minipage}\hfill
    \begin{minipage}{0.33\textwidth}
        \centering
        \includegraphics[width=\linewidth, height=5cm]{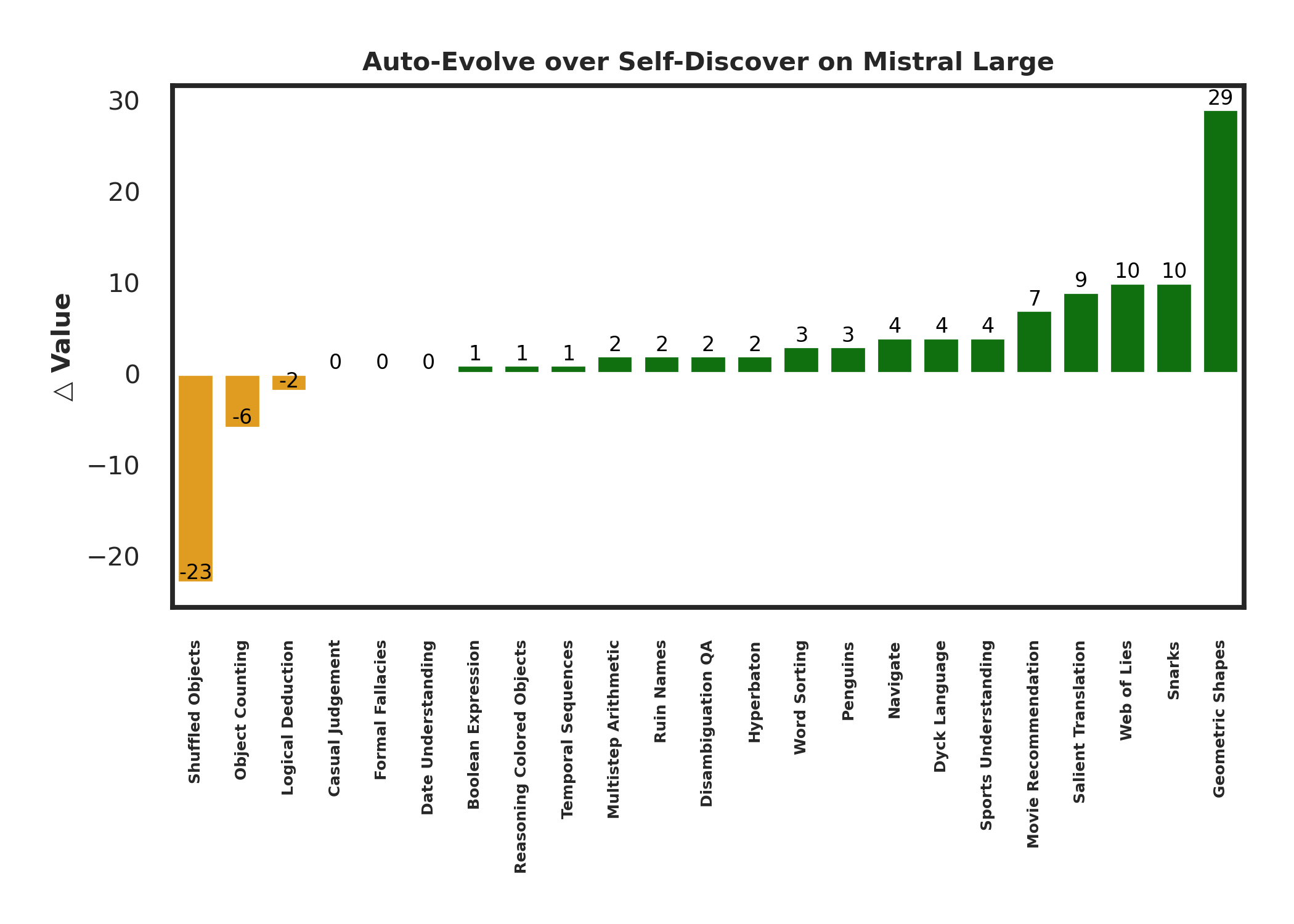}
    \end{minipage}
    \caption{
        Task level BBH performance on Mistral Large for Auto-Evolve over Direct Prompt, CoT and Self-Discover. Claude models and GPT-4 results are in Appendix \cref{fig:acc_diff_gpt} and \cref{fig:acc_diff_Claude_2.0}}
    \label{fig:acc_diff_mistral}
\end{figure*}

\subsection{Efficiency}

Auto-Evolve Framework is designed with efficiency and inference call costs in mind. For each task, the framework requires 1 call for \textbf{GENERATE}
, 1 call for \textbf{IMPLEMENT}
and on average 4-5 calls for \textbf{REFINE}. 
These one-time calls enable efficient processing of large datasets, with only 1 call per data point required once the reasoning structure is defined. Appendix \cref{fig:inference call} compares the efficiency of Auto-Evolve with other prompting framework (data from \citep{zhou2024selfdiscover}), demonstrating that it achieves similar or better performance than Self-Consistency and Majority Voting while requiring 10-40 times fewer inference calls.

\subsection{Themes: Improvement across categories}
\begin{figure}[!htbp]
\centering     \includegraphics[width=0.9\columnwidth]{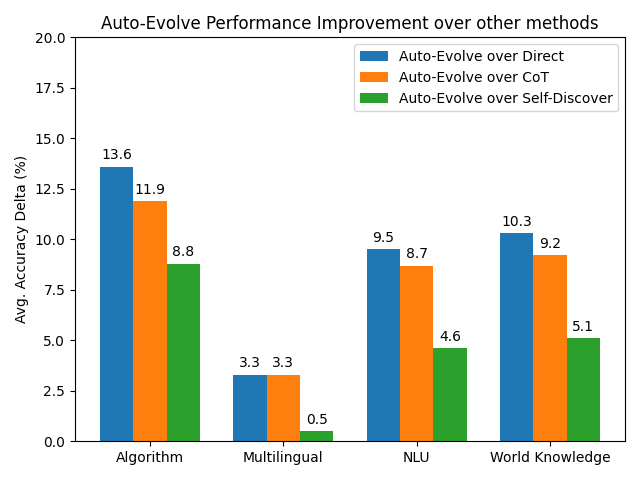}
      \caption{Performance of Auto-Evolve on Claude 2.0 in four task categories }
      \label{fig:categories_performance}
\end{figure}
Auto-Evolve demonstrates performance improvements across all four categories of the BBH dataset \citep{suzgun2022challenging} as shown in \cref{fig:categories_performance}. The most notable improvements are observed in the Algorithm category, where the complex reasoning structures generated by the Auto-Evolve prove particularly effective. We believe these types of tasks require much more complex reasoning structures because of which our framework outperformed Self-Discover.

\subsection{Ablation}

\begin{figure}[!h]
\centering      
 \includegraphics[width=0.9\columnwidth]{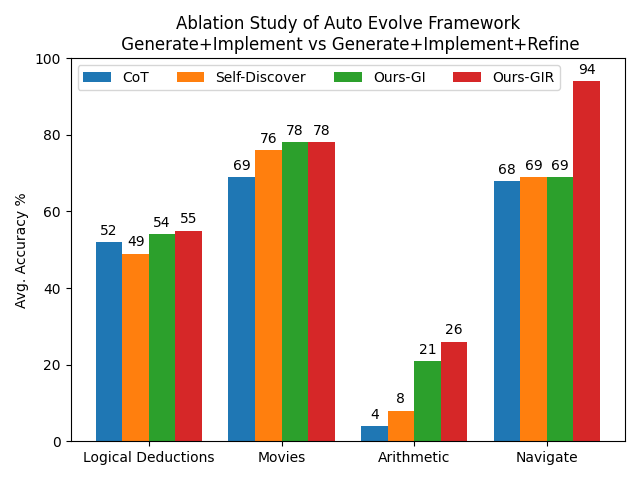}
      \caption{Auto-Evolve with and without REFINE on 4 diverse tasks on Claude 2.0}
      \label{fig:ablation_study}
\end{figure}

The ablation study in \cref{fig:ablation_study} highlights the individual contributions of the \textbf{GENERATE + IMPLEMENT} and \textbf{REFINE} components in the Auto-Evolve framework. These results are compared to the CoT and Self-Discover across fours tasks with Claude 2.0. We chose to conduct ablation study using Claude 2.0 as it had the most pronounced difference between results for Self-Discover and Auto-Evolve across the evaluated tasks (6.7\%), allowing us to clearly highlight the individual impacts of Auto-Evolve's components.

\textbf{GENERATE + IMPLEMENT} components alone for all the BBH tasks achieve 62.6\% performance. While with the refine step included it achieves 65.4\% performance, giving a performance boost of 2.8\%. It outperforms CoT and Self-Discover on all four tasks with avg. improvement of  7.25\% for CoT and 4.75\% for Self-Discover. With \textbf{GENERATE + IMPLEMENT} we see the most improvement in arithmetic task, 17\% on CoT, 13\% on Self-Discover. \textbf{REFINE} gives an avg. boost of 15\% for CoT and 12.75\% for Self-Discover.

In our experience, Auto-Evolve reasoning structures tend to increase in complexity in \textbf{REFINE} due to the cyclic incorporation of insights from multiple reasoning modules. While this complexity elevates performance in tasks demanding elaborate reasoning—such as Navigate, Arithmetic, and Date Understanding, it's not universally necessary. For the majority of tasks, \textbf{GENERATE + IMPLEMENT} contribute significantly to performance enhancements, achieving simpler yet efficient reasoning structures. \textbf{REFINE} should be selectively applied to complex tasks that demand deeper and more intricate reasoning capabilities.

\subsection{Deep Dive Analysis}
\subsubsection{Deep Diving into Auto-Evolve Reasoning Modules}

\cref{fig:deep_dive_reasoning_module} in Appendix showcases reasoning modules generated by Claude 2.0 using the Self-Discover and Auto-Evolve frameworks for two distinct tasks: Boolean and Disambiguation QA. In the case of the Boolean Expressions task, Auto-Evolve generates a highly pertinent module: "Identify and understand logical operators (not, and, or, etc.)", which directly addresses the core aspects of the task. On the other hand, Self-Discover uses more generic modules such as "Critical Thinking" and "Let's think step by step", which lack the task-specific focus needed for optimal performance.
Similarly, for the Disambiguation QA task, Auto-Evolve generates a module that captures the essence of the task: "Memory Module: Maintain awareness of noun phrases mentioned earlier in the passage or conversation to determine if the pronoun refers back to one of those". This module encapsulates the key aspects of pronoun resolution and antecedent identification, which are crucial for disambiguating references in the given context. In contrast, Self-Discover's modules remain more general, even after the adapt stage, where they are refined to "Identify the pronoun. Find all possible antecedents based on noun phrases". While this refinement improves the relevance of the modules, they still lack the complexity and specificity offered by Auto-Evolve. The enhanced relevance and specificity of Auto-Evolve's reasoning modules can be attributed to its ability to dynamically generate task-specific modules without relying on a fixed set of predefined seed modules.

\subsubsection{Deep Diving into Auto-Evolve Reasoning Structures}
In Appendix \cref{fig:deep_dive_analysis}, we showcase Auto-Evolve generated reasoning structures for Hyperbaton reasoning task using GPT-4. Auto-Evolve reasoning structure is tailored to its task, incorporating task-specific reasoning modules such as "Linguistic Analysis", "Adjective Order Rules", "Recall rules and examples" and etc. Through Linguistic Analysis reasoning module, Auto-Evolve is able to recognizes the standard English conventional order of adjectives, and derives the correct answer. Additionally, Appendix \cref{fig:deep_dive_analysis} contrasts reasoning processes from Self-Discover. Self-Discover's reasoning modules emphasize simplification and decomposition of problems into manageable parts, as well as consideration of human behavior nuances. Correspondingly, the generated reasoning structure breaks down sentences into constituent adjectives and simplifies grammatical rules to facilitate understanding. While Self-Discover also presents an action plan, it fails to recognize the task requirement of adjective ordering and yield incorrect answer. 

\subsection{Transferability and Generalizability to OpenSource Models}

One of the main challenges in using open-source models is achieving the same reasoning ability and accuracy as larger proprietary models. In our experiments on the disambiguation question-answering task from the BBH dataset, Llama 3.1 70B achieved only 22.4\% accuracy with direct prompting. However, with Auto-Evolve, which dynamically generates reasoning structures, the accuracy surged to 72.0\%, outperforming Self-Discover (56.8\%) and CoT (60.4\%) as well. We observed similar improvements on the causal judgement task, where Auto-Evolve (65.3\%) outperformed direct prompting (27.3\%), CoT (62.6\%), and Self-Discover (64.7\%).

Smaller models like Llama 3.1 8B typically struggle to generate complex reasoning plan autonomously. This limitation can be addressed by using larger models to create these reasoning structures. When we applied reasoning structures generated by Llama 3.1 70B to Llama 3.1 8B, the model's accuracy improved significantly. With Auto-Evolve, Llama 3.1 8B achieved 62.4\% accuracy compared to 45.2\% with direct prompting and 54.4\% with CoT. For the causal judgement task, Auto-Evolve (58.3\%)
again outperformed direct prompting (49.7\%) and
CoT (56.1\%).

These results highlight that while smaller models struggle to generate complex reasoning structures independently, they can perform well when guided by reasoning structures from larger models, demonstrating the transferability of reasoning strategies across model architectures. This approach provides an efficient solution for resource-constrained environments while still benefiting from advanced reasoning capabilities. It also opens opportunities for further research on optimizing transferability and balancing performance and efficiency across models of different sizes.

\section{Conclusion and Future Work}
Auto-Evolve introduces a novel framework that dynamically generates task-specific reasoning structures, eliminating the need for static seed modules and enabling more effective reasoning across diverse problem domains. By seamlessly integrating dynamic prompt generation and iterative refinement, Auto-Evolve surpasses state-of-the-art methods like CoT prompting, achieving performance improvements up to 10.4\% and an average gain of 6.8\% when evaluated with GPT-4, Claude 2.0, Claude 3 Sonnet and Mistral Large models. The framework's ability to transfer reasoning structures from larger models to smaller ones, as demonstrated with models like Llama 3.1 8B, highlights its broader utility and adaptability across architectures.

The broader implications of Auto-Evolve extend beyond the performance enhancement, as the framework has the potential to advance the development of more interpretable and transparent AI systems by generating dynamic problem specific reasoning modules and explicit reasoning structures. Our experimentation has highlighted the pivotal role played by JSON reasoning structures in solving tasks effectively. In future iterations, we aim to explore the potential of incorporating feedback mechanisms to iteratively improve these reasoning structures, further refining and enhancing the framework's capabilities.

\section*{Limitations}
While the proposed Auto-Evolve framework demonstrates promising results in enhancing large language models' reasoning capabilities, we acknowledge the following limitations:

Applicability to Smaller Models: Our experiments demonstrate that large models like Llama 3.1 70B can directly benefit from Auto-Evolve, independently generating and utilizing sophisticated reasoning structures. However, smaller models such as Llama 3.1 8B struggle to create these structures autonomously. We found that applying reasoning structures generated by larger models (e.g., Llama 3.1 70B) to guide smaller models significantly enhances their performance. This combined approach enables resource-efficient models to leverage advanced reasoning capabilities. Future research will focus on optimizing this transfer process, exploring methods to effectively scale reasoning capabilities across models of varying sizes and architectures, with particular emphasis on enhancing smaller, more efficient models using insights from their larger counterparts.

Increased Complexity in Reasoning Structures: Auto-Evolve’s reasoning structures can become overly complex due to the cyclic incorporation of insights from multiple reasoning modules. This complexity, while beneficial for certain tasks demanding elaborate reasoning, is not universally necessary and can be an overhead for simpler tasks. Based on our experience we suggest readers to incorporate all reasoning modules in a single step as a starting point and then use iterative part of the framework as a optional step for problems that can't be solved with single step.

Model Determinism: During our experiments, we observed non-deterministic behavior even when the temperature was set to be 0. Slight variations in the generated reasoning modules led to significant disparities in the downstream reasoning structures and outputs. To address this, we ran multiple trials and reported average performance, which added computational overhead.

\section*{Ethics Statement}

Bias Propagation and Amplification: While Auto-Evolve is designed to enhance the reasoning abilities of large language models (LLMs), we acknowledge the potential for the generated reasoning modules to propagate or even amplify biases present in the underlying model's training data. If the training data contains cultural, societal, or linguistic biases, these biases may manifest in the reasoning modules and structures produced by Auto-Evolve. To mitigate this risk, it is crucial to incorporate human-in-the-loop feedback mechanisms or other guardrails to ensure that the final outputs align with user values and ethical considerations.

\section*{Acknowledgements}
We would like to thank Callin Switzer, Jane Barker and Kai Wei for their thorough review of this paper and feedback. We also appreciate Greg Sansoni and Stephanie Kim for their support.

\bibliography{auto_evolve}
\bibliographystyle{acl_natbib}

\onecolumn
\appendix

\section*{Appendix}
\label{sec:appendix}

\section{Auto-Evolve Prompt details}

\begin{figure*}[h]
  \includegraphics[width=\linewidth,height=8cm]{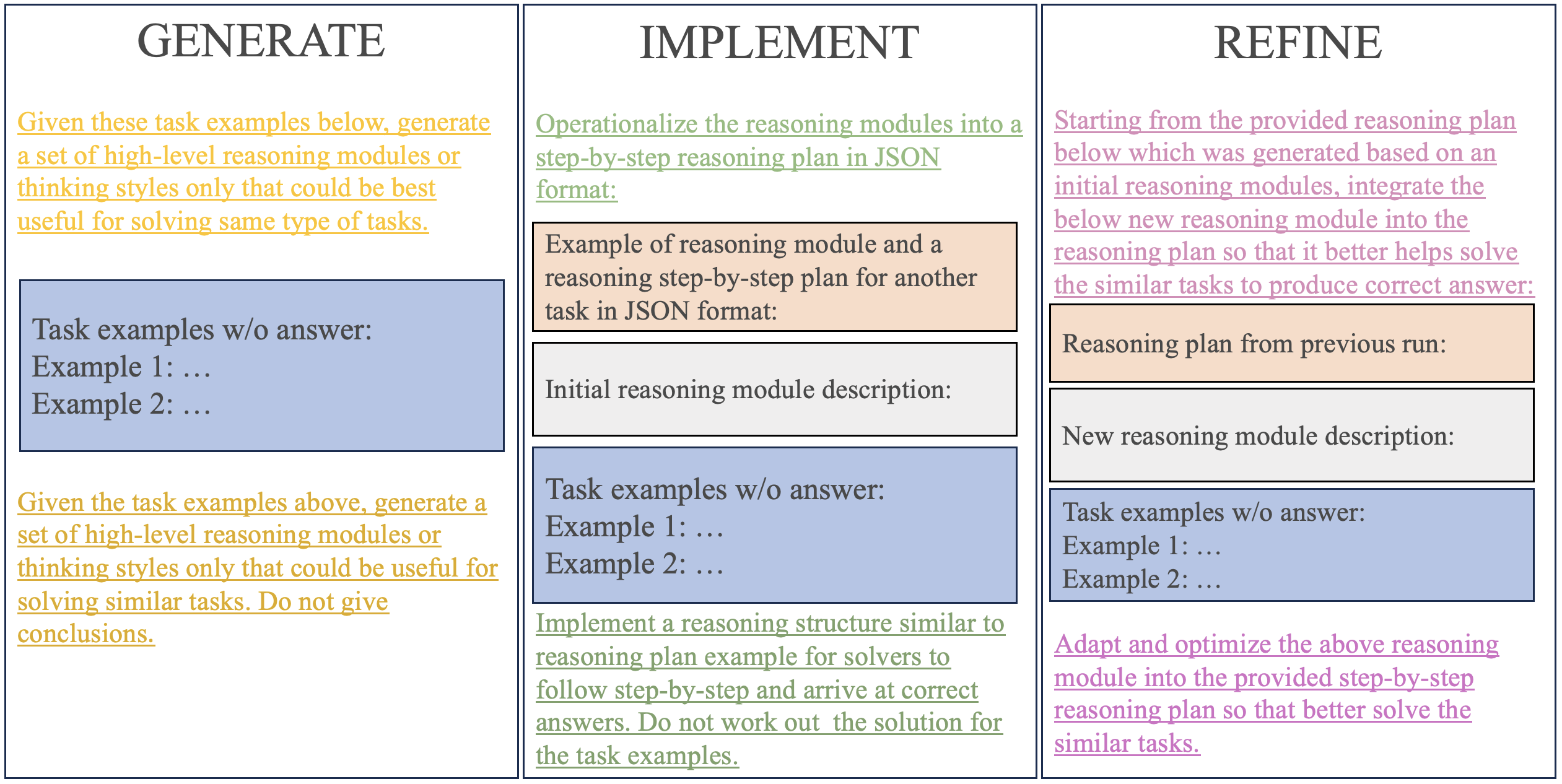}
  \caption{Meta-Prompts for the three components of Auto-Evolve. }
  \label{fig:self-evolve promt detail}
\end{figure*}

The meta-prompt templates for the \textbf{GENERATE}, \textbf{IMPLEMENT} and \textbf{REFINE} components in the first stage of Auto-Evolve are shown in \cref{fig:self-evolve promt detail}.

\section{Performance on BBH dataset}
\cref{table:bbh_pertask} contains BBH per-task performance of Claude 2.0, Claude 3 Sonnet and Mistral Large over 4 prompt strategies comparing to human performance. Compared to human average performance, Mistral Large with Auto-Evolve framework outperforms on 19 out of 23 tasks, Claude 2.0 with Auto-Evolve outperforms on 11 out of 23 tasks, and Claude 3 Sonnet with Auto-Framework outperforms on 11 out of 23 tasks.

\begin{table*}[!htbp]
\caption{Big Bench-Hard~\citep{suzgun2022challenging} per-task performance of Claude 2.0, Claude 3 Sonnet and Mistral Large with Auto-Evolve, the highest accuracy for each task has been highlighted in bold.}
\medskip
\centering
\resizebox{\linewidth}{!}{
\begin{tabular}{l|cc|cccc|cccc|cccc}
\toprule
\textbf{Big Bench-Hard Task}&  \textbf{\begin{tabular}[c]{@{}c@{}}Human\\ (Avg.)\end{tabular}}&  \textbf{\begin{tabular}[c]{@{}c@{}}Human\\ (Max)\end{tabular}}& \textbf{\begin{tabular}[c]{@{}c@{}}Mistral-L\\ Direct\end{tabular}}&  \textbf{\begin{tabular}[c]{@{}c@{}}Mistral-L\\ + CoT\end{tabular}}&  \textbf{\begin{tabular}[c]{@{}c@{}}Mistral-L\\ + Self-Discover\end{tabular}}&  \textbf{\begin{tabular}[c]{@{}c@{}}Mistral-L\\ + Auto-Evolve\end{tabular}}&  \textbf{\begin{tabular}[c]{@{}c@{}}Claude 2.0\\ Direct\end{tabular}}& \textbf{\begin{tabular}[c]{@{}c@{}}Claude 2.0\\ + CoT\end{tabular}}&   \textbf{\begin{tabular}[c]{@{}c@{}}Claude 2.0\\ + Self-Discover\end{tabular}}&\textbf{\begin{tabular}[c]{@{}c@{}}Claude 2.0\\ + Auto-Evolve\end{tabular}}&  \textbf{\begin{tabular}[c]{@{}c@{}}Claude 3 Sonnet\\ Direct\end{tabular}}& \textbf{\begin{tabular}[c]{@{}c@{}}Claude 3 Sonnet\\ + CoT\end{tabular}}&   \textbf{\begin{tabular}[c]{@{}c@{}}Claude 3 Sonnet\\ + Self-Discover\end{tabular}}&\textbf{\begin{tabular}[c]{@{}c@{}}Claude 3 Sonnet\\ + Auto-Evolve\end{tabular}}\\

\midrule
boolean\_expressions& 79& 100& 75& 79& 92& \textbf{93}& 79& 79& 78&\textbf{86}& 94& \textbf{98}& 93& 90\\
causal\_judgement&  70&  100&  67&  72&  72& \textbf{72}&   61&  61&  65&\textbf{67}& \textbf{69}& 68& 58& 67\\
date\_understanding&  77&  100&  67&  69&  79&  \textbf{80}&  53&  56&  \textbf{72}&70& 66& 66& 65& \textbf{74}\\
disambiguation\_qa&  67&  93&  67&  65&  76&  \textbf{77}&  58&  60&  \textbf{70}&68& 54& 52& 68& \textbf{70}\\
dyck\_languages&  48&  100&  20&  \textbf{23}&  10&  14&  14&  \textbf{14}&  13&10& 11& 8& 16& \textbf{19}\\
formal\_fallacies&  91&  100&  53&  53&  53&  \textbf{53}&  53&  53&  53&\textbf{53}& 53& 53& 58& \textbf{59}\\
geometric\_shapes&  54&  100&  28&  31&  38&  \textbf{67}&  38&  39&  37&\textbf{49}& 47& 44& 51& \textbf{66}\\
hyperbaton&  75&  100&  82&  81&  88&  \textbf{89}&  62&  64&  64&\textbf{76}& 72& 73& \textbf{81}& 70\\
logical\_deduction\_seven\_objects&  40&  89&  57&  60&  \textbf{65}&  62&  52&  52&  49&\textbf{55}& 56& 56& \textbf{62}& 56\\
 movie\_recommendation& 61& 90& 75& 74& 74& \textbf{80}& 68& 69& 76&\textbf{78}& 75& 75& \textbf{84}& 83\\
 multistep\_arithmetic\_two&  10&  25&  20&  \textbf{60}&  55&  57&  3&  4&  8&\textbf{26}& \textbf{73}& 71& 56& 60\\
 navigate& 82& 100& 73& 73& 85& \textbf{88}& 48& 68& 69&\textbf{94}& 62& 74& \textbf{88}& 86\\
 object\_counting& 86& 100& 58& 65& \textbf{80}& 74& 52& 53& 54&\textbf{60}& 74& \textbf{79}& 76& 76\\
 penguins\_in\_a\_table& 78& 100& 61& 68& 83& \textbf{86}& 57& 60& 69&\textbf{78}& 75& 80& \textbf{82}& 74\\
 reasoning\_about\_colored\_objects& 75& 100& 79& 82& 83& \textbf{84}& 59& 61& 68&\textbf{76}& 79& 76& 79& \textbf{82}\\
 ruin\_names& 78& 100& 78& 79& 81& \textbf{83}& 61& 60& 54&\textbf{71}& 71& 70& 72& \textbf{76}\\
 salient\_translation\_error\_detection& 37& 80& 58& 59& 60& \textbf{69}& 58& 58& 61&\textbf{61}& 65& 65& 64& \textbf{68}\\
 snarks& 77& 100& 75& 77& 77& \textbf{87}&  69& 67& 66&\textbf{71}& 70& \textbf{72}& 70& 70\\
 sports\_understanding& 71& 100& 79& 80& 81& \textbf{85}& 71& 73& 74&\textbf{79}& 76& 78& 70& \textbf{85}\\
 temporal\_sequences& 91& 100& 93& 94& 98& \textbf{99}& 62& 60& 65&\textbf{73}& 92& 84& \textbf{95}& 90\\
 tracking\_shuffled\_objects\_seven\_objects& 65& 100& 22& 66& \textbf{72}& 49& 18& 16& 43&\textbf{51}& \textbf{90}& 72& 37& 64\\
 web\_of\_lies& 81& 100& 50& 51& 83& \textbf{93}& 49& 48& 52&\textbf{62}& \textbf{77}& 74& 49& 67\\
 word\_sorting& 63& 100& 88& 87& 89& \textbf{92}& \textbf{91}& 90& 90&90& 77& 81& 82& \textbf{94}\\
 \bottomrule
\end{tabular}
}
\label{table:bbh_pertask}
\end{table*}


\section{Analyzing Reasoning Processes}
The comparison between reasoning modules generated using Self-Discover and Auto-Evolve reveals distinct approaches to problem-solving. Self-Discover's generated reasoning modules emphasize simplification and decomposition of problems into manageable parts, as well as consideration of human behavior nuances. Correspondingly, the generated reasoning structure breaks down sentences into constituent adjectives and simplifies grammatical rules to facilitate understanding. In contrast, Auto-Evolve's task-specific reasoning modules prioritize linguistic and critical analysis for evaluating sentence structures. The resulting reasoning structure involves pattern and comparative analyses to identify adherence to standard adjective order rules. Ultimately, Auto-Evolve's approach yields the correct answer by systematically analyzing sentence structures and identifying deviations from conventional rules, showcasing its effectiveness in task-specific problem-solving.
\begin{figure*}[!htbp]
  \centering
  \includegraphics[width=\textwidth,height=8cm]{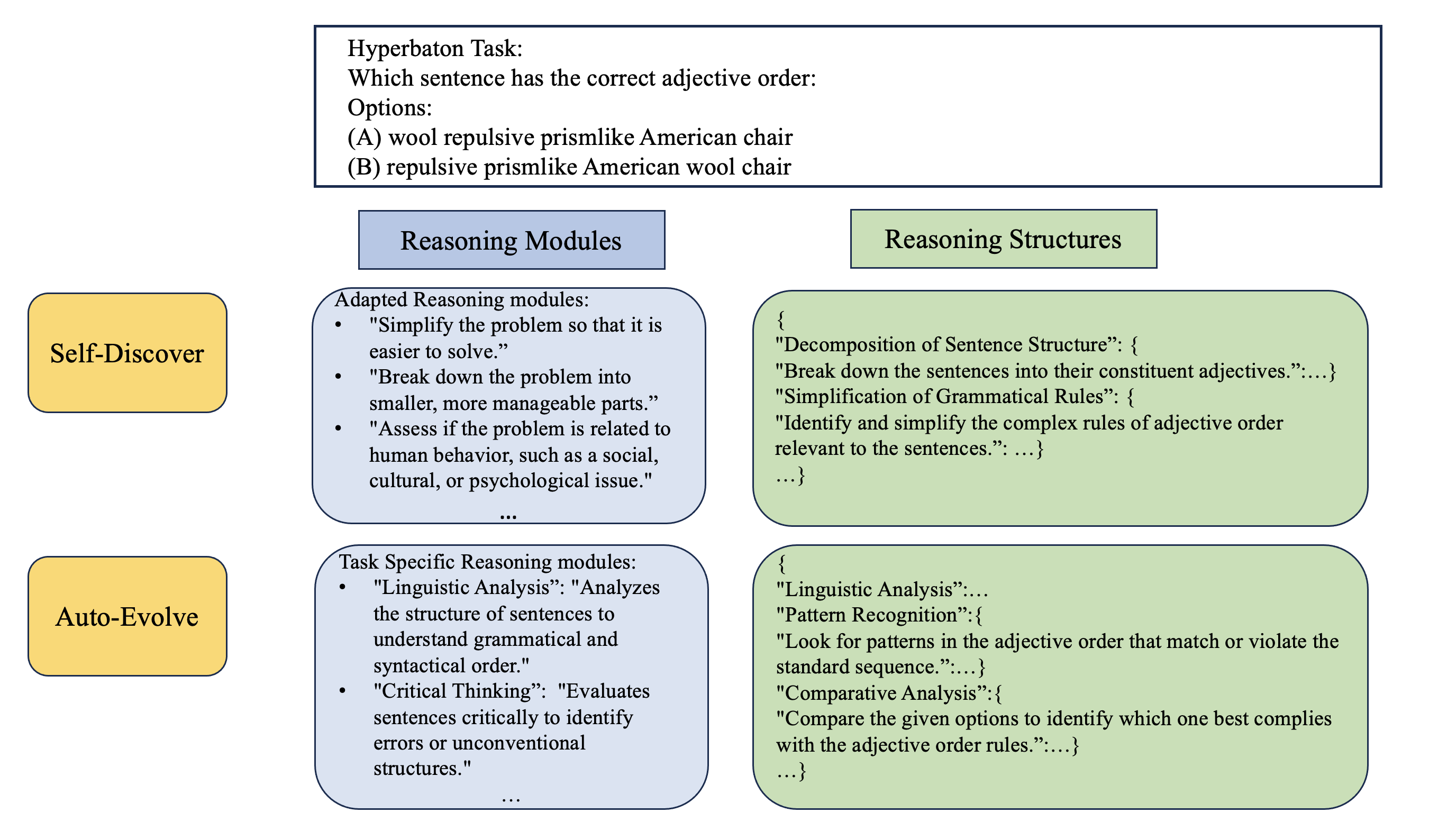}
  \caption{Comparison between Auto-Evolve and Self-Discover reasoning modules and reasoning structure process generated from GPT-4 on a Hyperbaton task.}
  \label{fig:self-evolve reasoning process}
\end{figure*}

\section{Reasoning Module Analysis}
\begin{figure*}[!htbp]
  \centering
  \includegraphics[width=16cm,height=8cm]{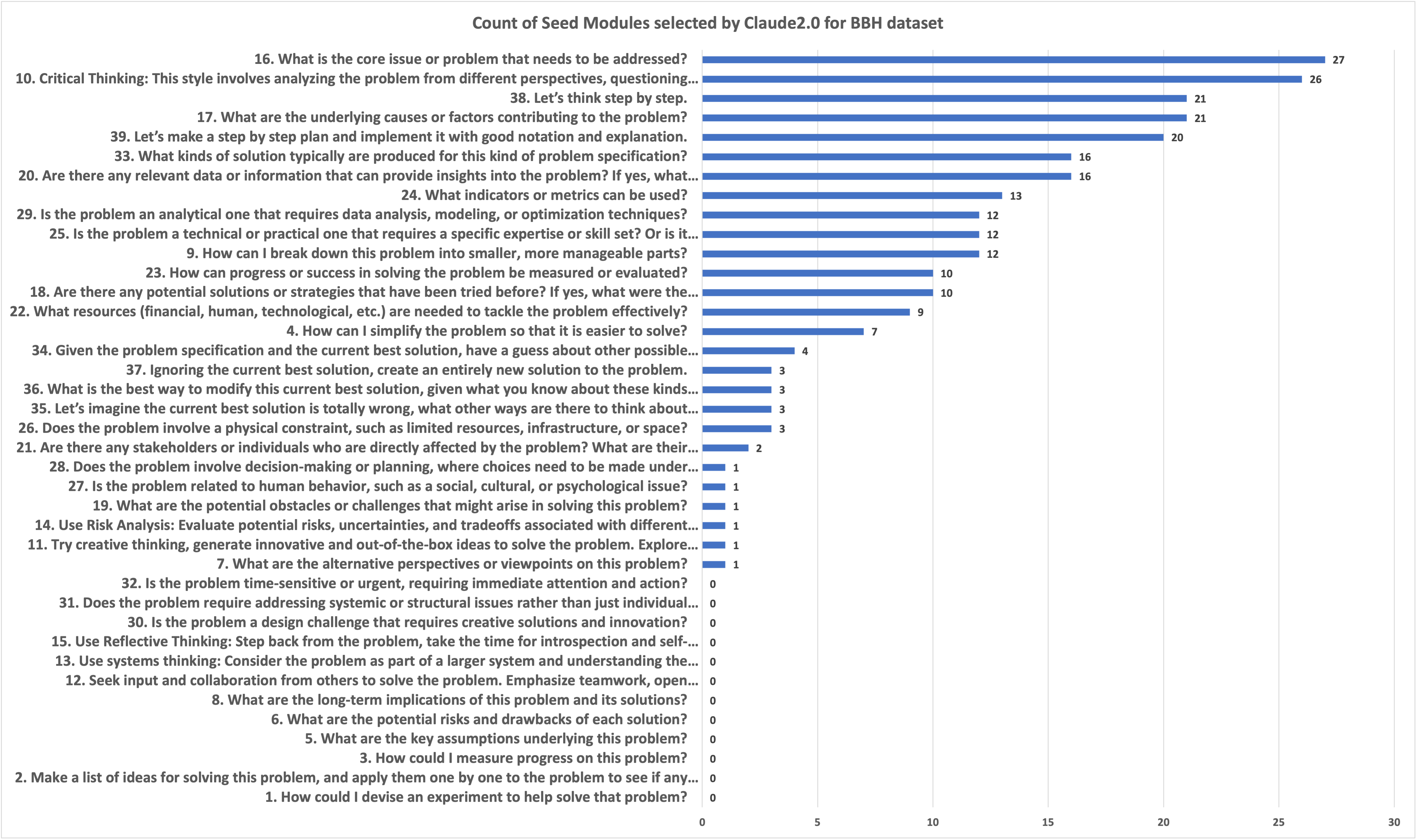}
  \caption{Analysis of Self-Discover seed modules and how these were selected by Claude2.0 for BBH dataset. }
  \label{fig:seed_modules_analysis}
\end{figure*}

Frequency plot \cref{fig:seed_modules_analysis} showcases that Self-Discover only uses a few reasoning seed modules in solving the BBH tasks (out of 39).
The inherent gravitation of LLMs towards utilizing only a subset of the provided seed reasoning modules can stem from a variety of factors. This tendency may arise due to inherent biases within the models, leading them to preferentially select familiar patterns or reasoning strategies. Alternatively, the lack of diversity within the seed module set itself, with many modules representing relatively similar reasoning approaches, could compel the model to gravitate towards a distinct few. We believe that using a fixed set of human-defined seed modules introduces inductive biases that constrain the model's reasoning flexibility across diverse tasks, compared to Auto-Evolve's approach of dynamically generating tailored reasoning modules for each task type.

\section{Auto-Evolve Reasoning Module Comparison}
\begin{figure*}[!htbp]
  \centering
  \includegraphics[width=\textwidth,height=8cm]{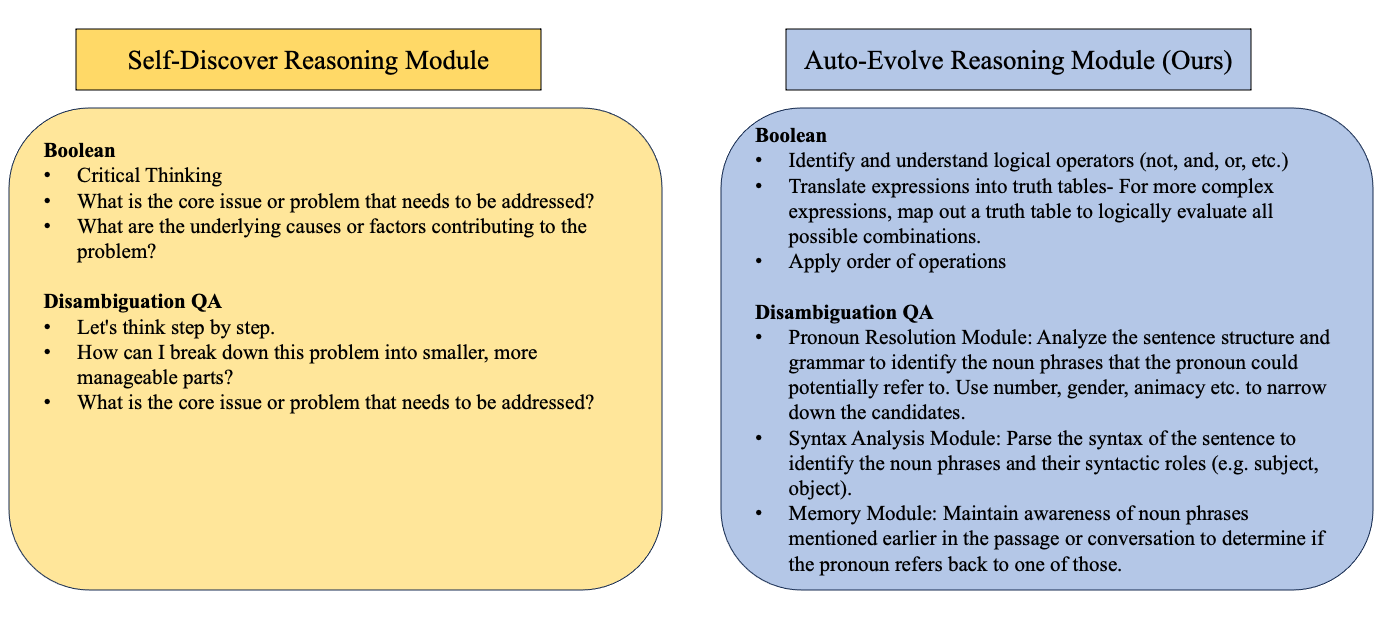}
  \caption{Deep Dive Analysis of Self-Discover vs Auto-Evolve Reasoning Module }
  \label{fig:deep_dive_reasoning_module}
\end{figure*}

\cref{fig:deep_dive_reasoning_module} shows deep analysis on reasoning module generation comparisons across two different prompt strategies (Self-Discover and Auto-Evolve). For both Boolean and Disambiguation tasks, Auto-Evolve represents a significant advancement over Self-Discover by implementing more detailed, task-specific reasoning modules. This approach allows for greater flexibility and adaptability, enhancing the model's performance in complex reasoning tasks. The specific focus on logical operations, detailed syntax and grammar analysis, and memory retention provides a more comprehensive framework for improving LLM reasoning capabilities.

\section{Auto-Evolve Reasoning Structure Comparison}

\begin{figure*}[!h]
  \centering
  \includegraphics[width=\textwidth,height=7cm]{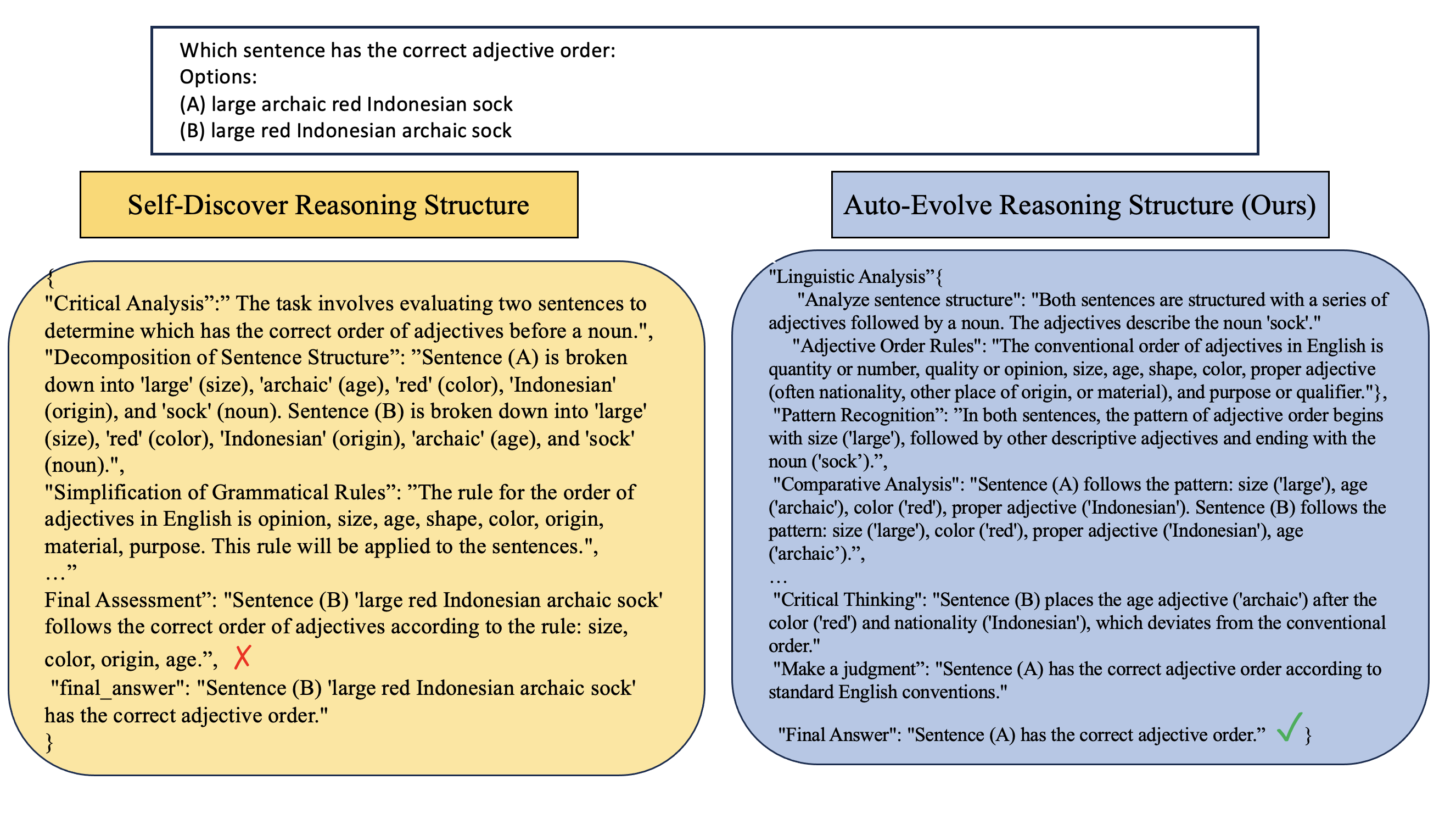}
  \caption{Deep Dive Analysis of Self-Discover vs Auto-Evolve Reasoning Structure }
  \label{fig:deep_dive_analysis}
\end{figure*}
In \cref{fig:deep_dive_analysis} example, it showcases the LLMs follow the reasoning structures using Auto-Evolve and Self-Discover framework on a Hyperbaton task. LLMs are able to follow the Auto-Evolve's guidance integrated with task-specific instructions (keys and sub-keys) and derive the final answer correctly.
In this specific Hyperbaton task, while Self-Discover relies on a broad and generic analysis, Auto-Evolve employs a detailed and structured approach that includes linguistic analysis, pattern recognition, and comparative evaluation. This comprehensive method allows Auto-Evolve to accurately apply grammatical rules and critically assess sentence structures, leading to more reliable and correct outcomes. The Auto-Evolve framework’s ability to dynamically adapt its reasoning structures based on the specific task at hand demonstrates a significant improvement in handling this complex linguistic challenges.
\newline
\newline
\newline
\newline

\section{Auto-Evolve Performance Comparison}
\label{Auto-Evolve Performance Comparison}
\begin{figure*}[!htbp]
    \centering
    \begin{minipage}{0.33\textwidth}
        \centering
        \includegraphics[height=5cm, width=\linewidth]{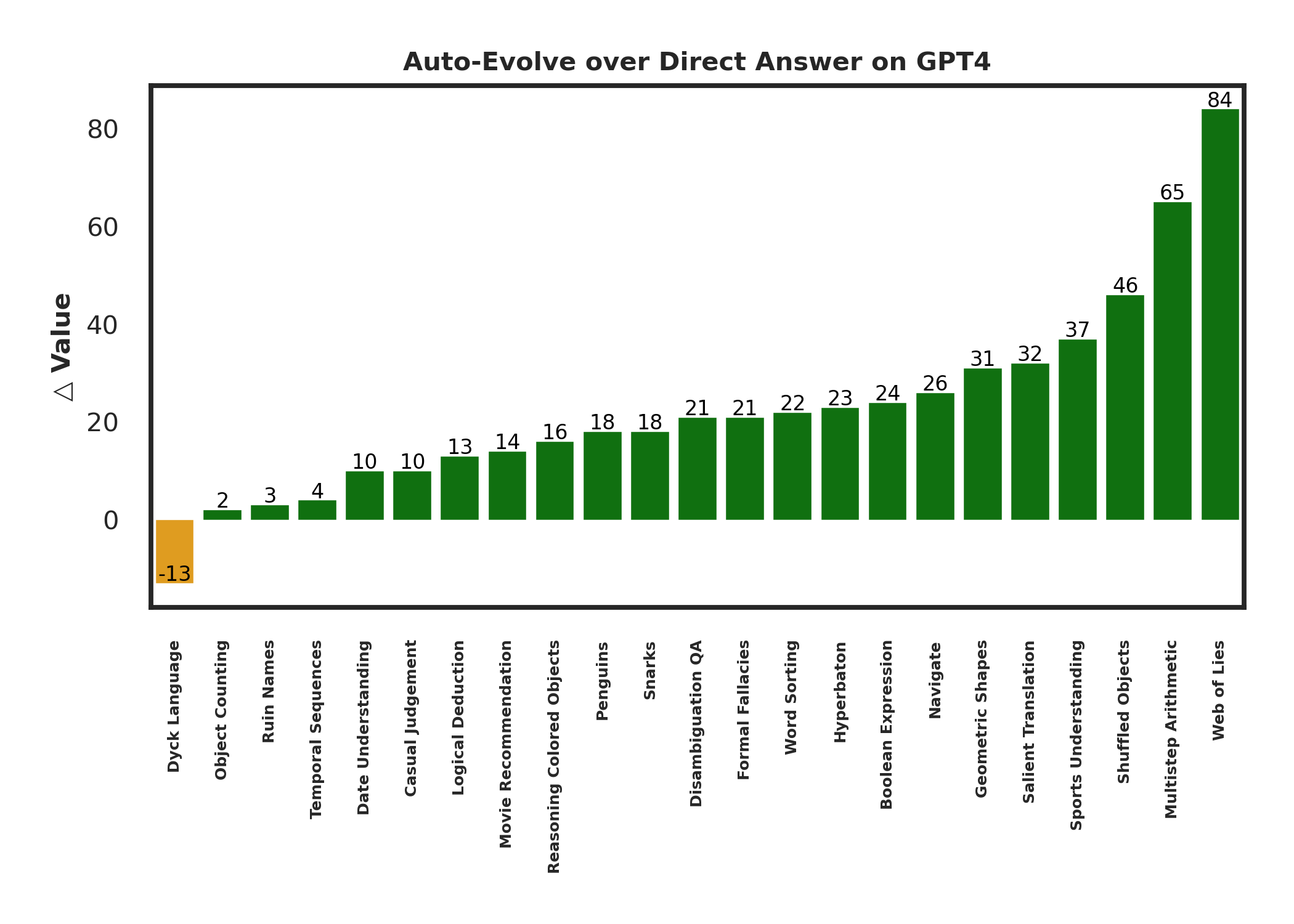}
    \end{minipage}
    \begin{minipage}{0.33\textwidth}
        \centering
        \includegraphics[height=5cm, width=\linewidth]{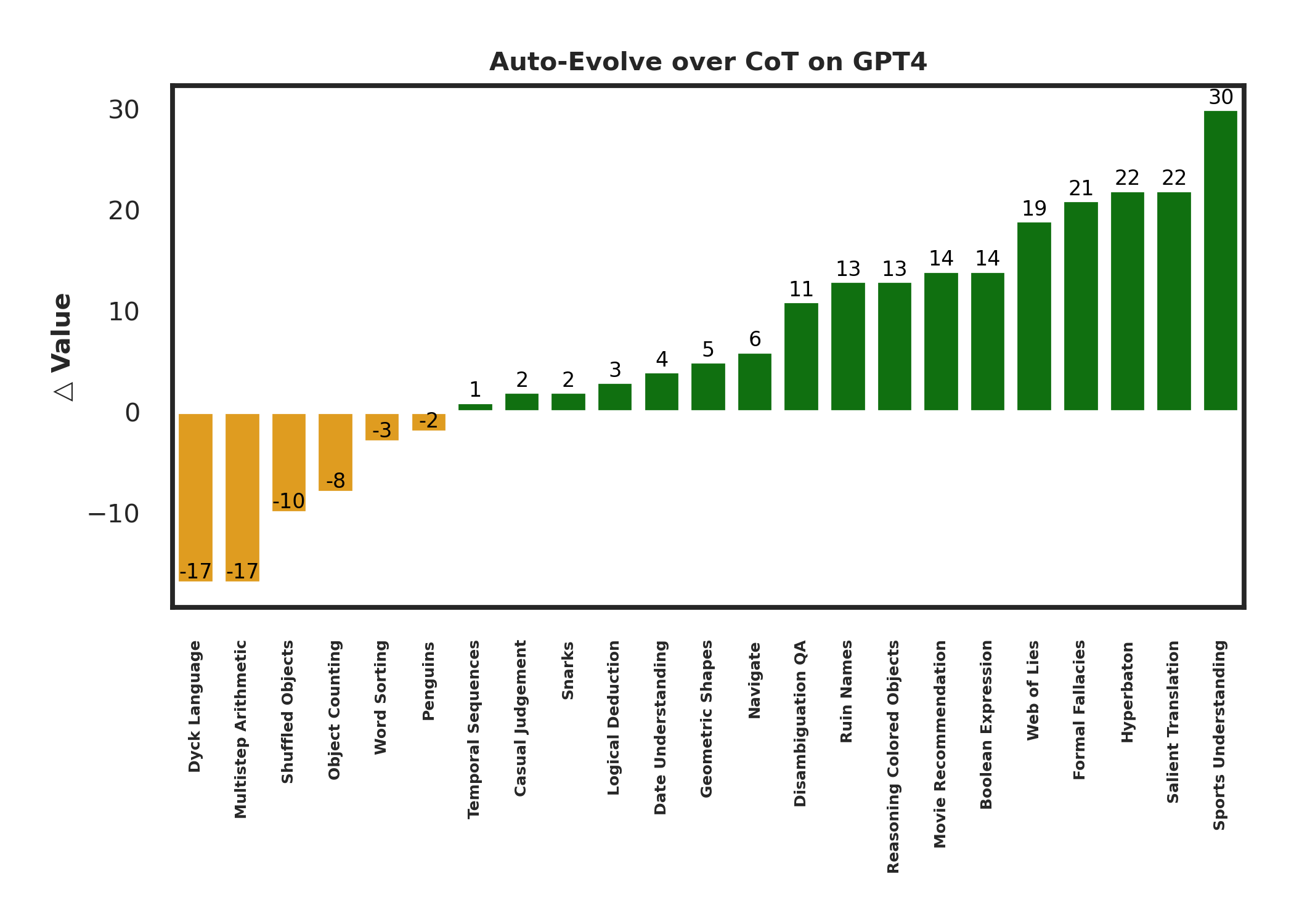}
    \end{minipage}\hfill
    \begin{minipage}{0.33\textwidth}
        \centering
        \includegraphics[height=5cm, width=\linewidth]{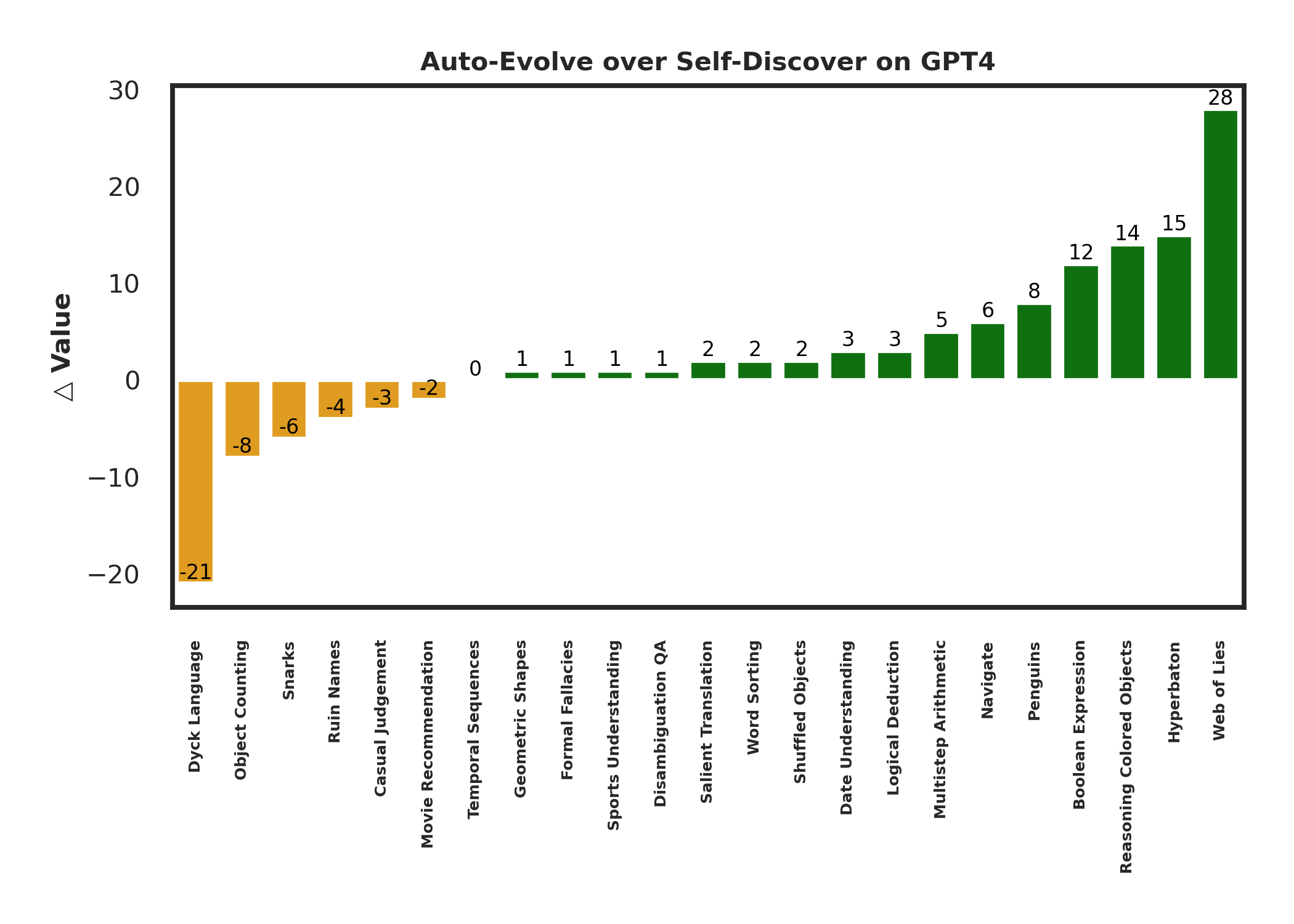}
    \end{minipage}
    \caption{Performance comparison between Auto-Evolve and Direct Prompt, CoT and Self-Discover on GPT-4. }
    \label{fig:acc_diff_gpt}
\end{figure*}

In the \cref{fig:acc_diff_gpt}, it displays the accuracy differences of Auto-Evolve over Direct Prompt, CoT and Self-Discover on GPT-4 for BBH 23 tasks. The green bars show the absolute percentage improvement, and yellow bars show the absolute percentage decrease. Auto-Evolve outperforms 22/23 tasks over Direct Prompt, and outperforms 17/23 tasks over CoT and Self-Discover on GPT-4. By using GPT-4 with our proposed framework Auto-Evolve, it improves most on complex tasks such as Web of Lies, Multistep Arithmetic, Shuffled Object and etc. 

\begin{figure*}[!htbp]
    \centering
    \begin{minipage}{0.33\textwidth}
        \centering
        \includegraphics[height=5cm, width=\linewidth]{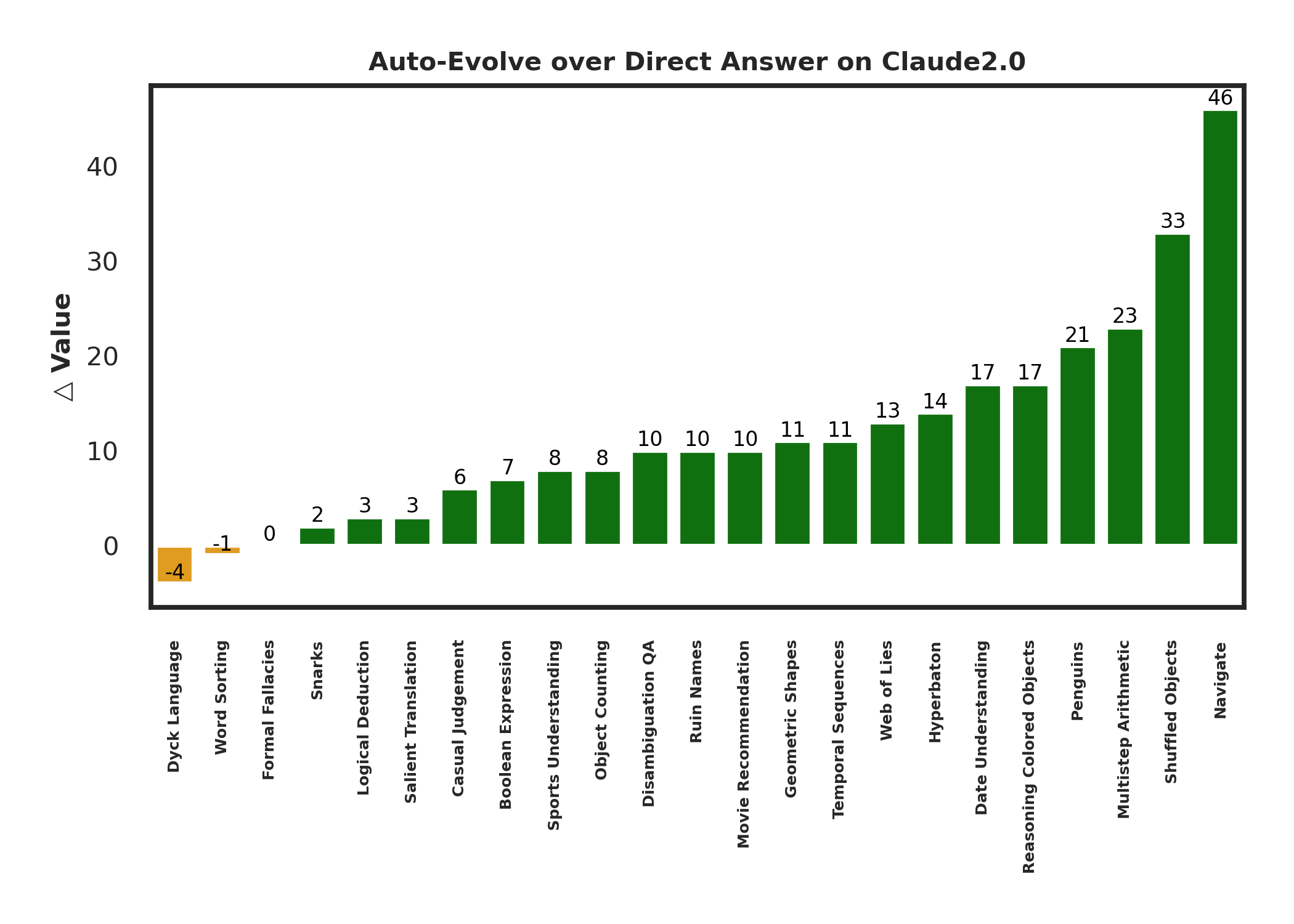}
    \end{minipage}
    \begin{minipage}{0.33\textwidth}
        \centering
        \includegraphics[height=5cm, width=\linewidth]{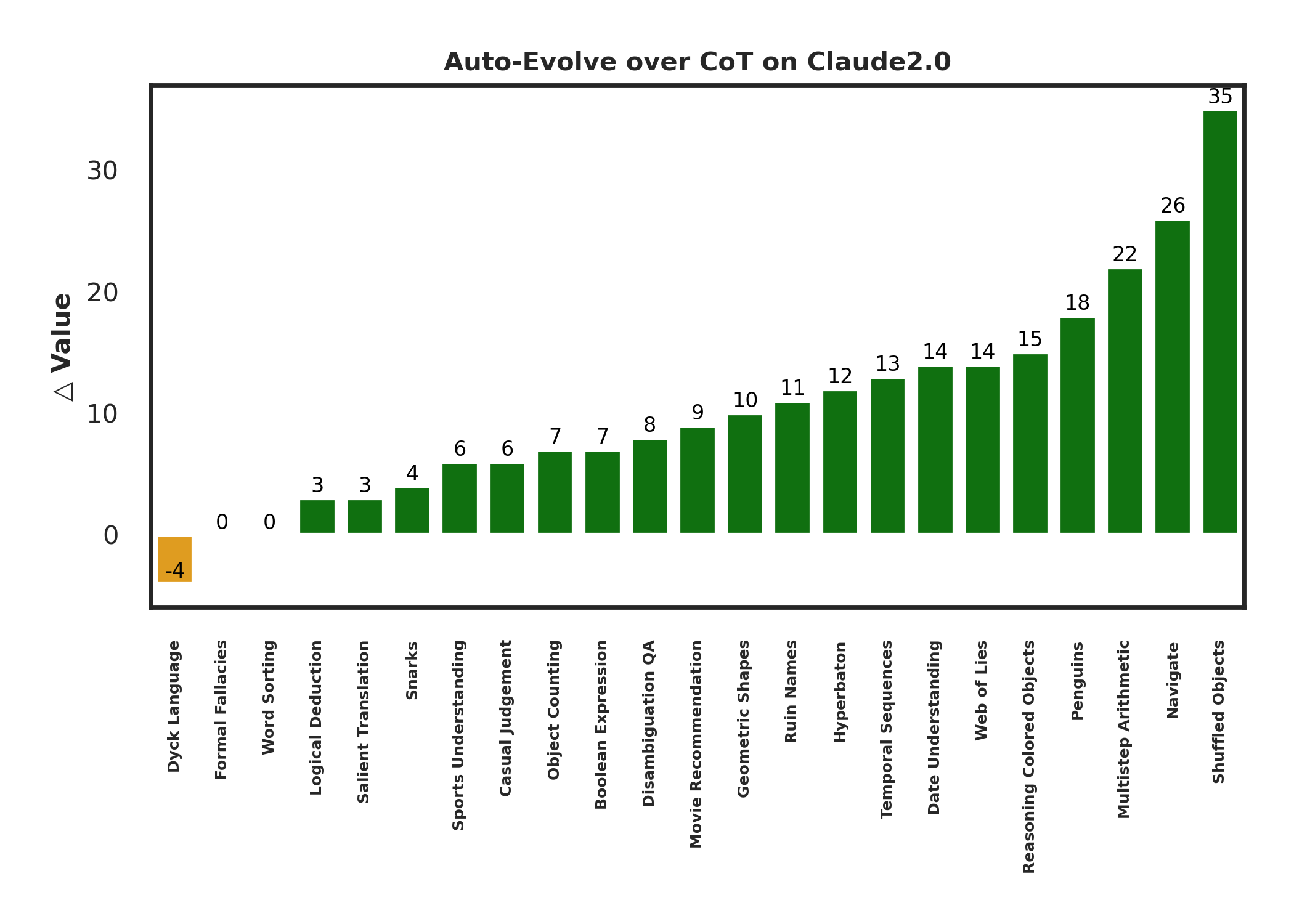}
    \end{minipage}\hfill
    \begin{minipage}{0.33\textwidth}
        \centering
        \includegraphics[height=5cm, width=\linewidth]{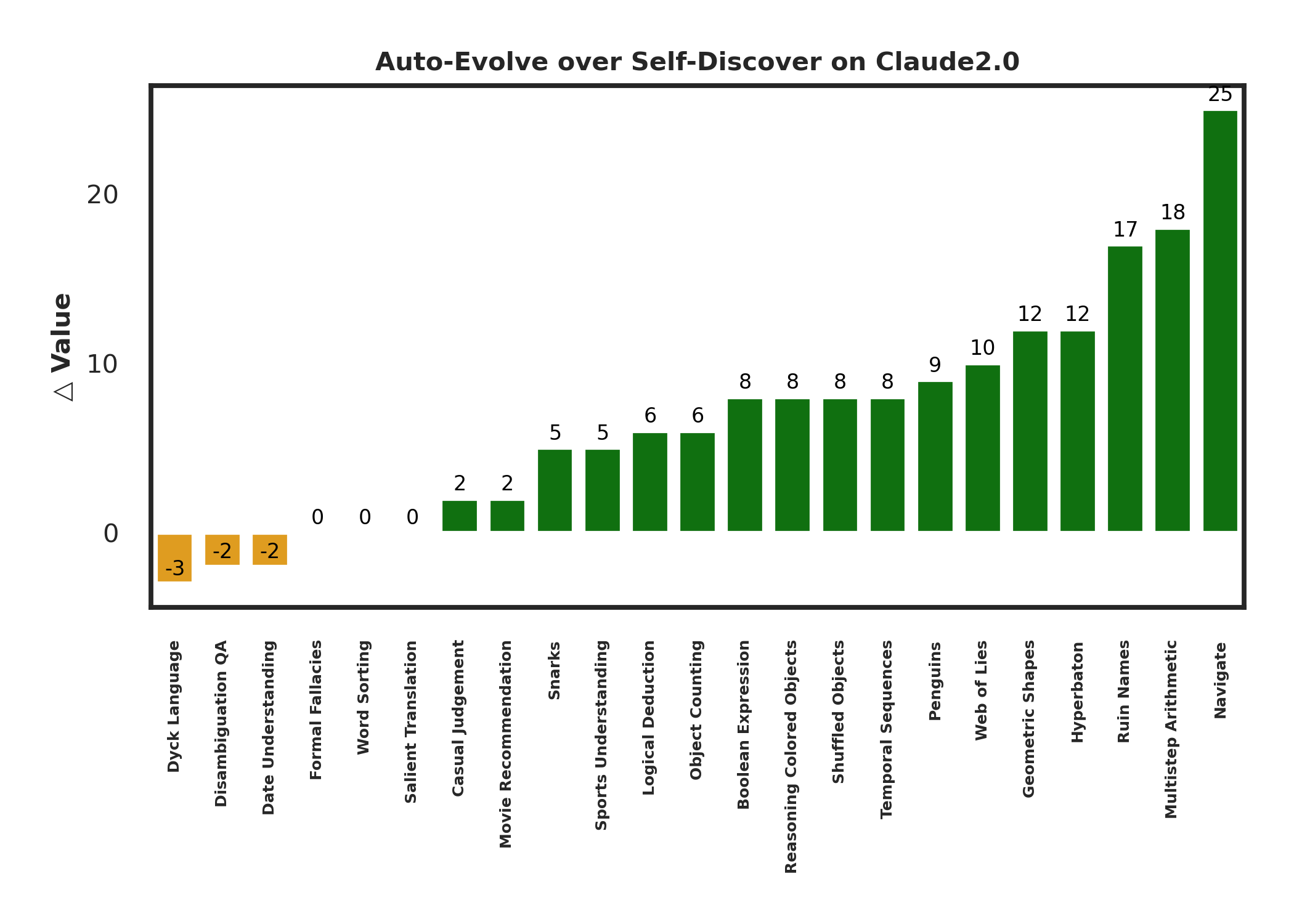}
    \end{minipage}
    \caption{
        Task level BBH performance on Claude 2.0 for Auto-Evolve over Direct Prompt, CoT and Self-Discover.}
    \label{fig:acc_diff_Claude_2.0}
\end{figure*}

In the \cref{fig:acc_diff_Claude_2.0}, it displays the accuracy differences of Auto-Evolve over Direct Prompt, CoT and Self-Discover on Claude 2.0 for 23 tasks. Auto-Evolve outperforms 20/23 tasks over Direct Prompt and CoT, and outperforms 17/23 tasks over Self-Discover. By using Claude 2.0 with our proposed framework Auto-Evolve, it improves most on complex tasks such as Navigate, Multistep Arithmetic, Shuffled Object and etc. 

\section{Efficiency Comparison}

\begin{figure*}[!h]
  \includegraphics[width=\textwidth,height=5cm]{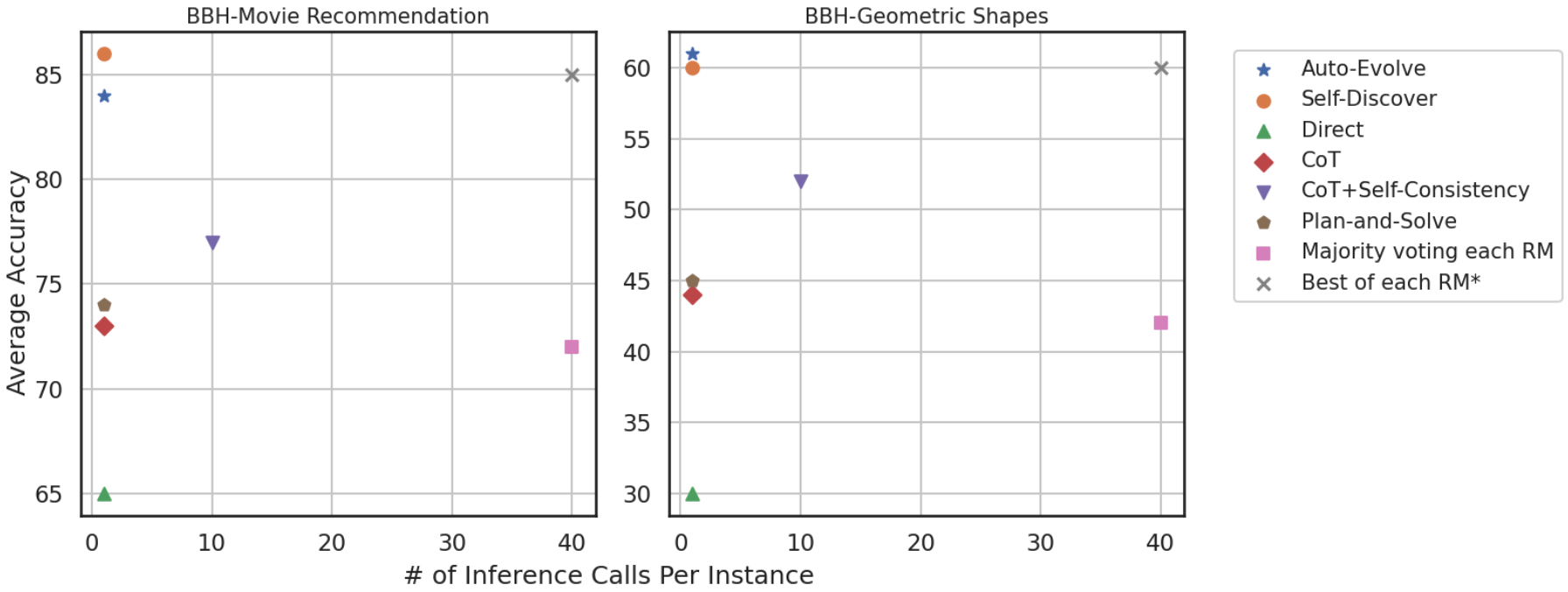}
  \caption{Number of inference calls vs average accuracy comparison on GPT-4 per task instance on Movie Recommendation task and Geometric Shapes task. We obtain the other data (Plan-and-Solve, etc) from \citep{zhou2024selfdiscover}. Auto-Evolve framework requires lowest number of inference calls per instance while maintain the highest or on-par performances on accuracy for Movie Recommendation and Geometric Shapes tasks  }
  \label{fig:inference call}
\end{figure*}
In \cref{fig:inference call}, it displays the number of inference calls per task instance. Below, we give an example of total number of calls by task level (aggregate level).
Example: For one task which includes \textbf{250} questions, below are the number of inference calls to the LLMs for different prompting strategies compared to Auto-Evolve.
\textbf{Direct Prompting}: 250 calls / per task.

\textbf{Chain-of-Thought}: 250 calls / per task.

\textbf{Self-Discover}: 3 calls (First Part meta prompt) + 1*250 instances = 253 calls / per task.

\textbf{Cot+Self-Consistency}: Sample 10 times, 10*250 instances = 2500 calls / per task.

\textbf{majority voting of each Reward Model}: Require golden labels, 40*250 instances = 10K calls / per task.

\textbf{Auto-Evolve}: 6$\sim$7 calls (Include iterative Refinement) + 1*250 instances $\approx$ 256  calls / per task.

\end{document}